\documentclass[10pt]{article}
\usepackage[utf8]{inputenc}
\usepackage[T1]{fontenc}
\usepackage[
backend=bibtex,
sorting=ynt
]{biblatex}
\usepackage{caption}
\usepackage{subcaption}
\usepackage{amsmath}
\usepackage{multirow}
\newcommand{\KL}{\mathrm{KL}}
\usepackage{PRIMEarxiv}
\usepackage{array}
\usepackage{hyperref}       
\usepackage{url}            
\usepackage{booktabs}       
\usepackage{amsfonts}       
\usepackage{wrapfig}
\usepackage{nicefrac}       
\usepackage{float}
\usepackage{tabularx}
\usepackage{microtype}      
\usepackage{lipsum}
\usepackage{listings}
\usepackage{xcolor} 
\usepackage{fancyhdr}       
\usepackage{graphicx} 
\usepackage{pythonhighlight}
\usepackage[most]{tcolorbox}

\title{

    \centering
    \fontsize{14}{16}\selectfont \textbf{Towards Evaluating Robustness of Prompt Adherence in Text to Image Models}\
    }
    
\vspace{1em}
\author{
    \begin{minipage}{0.5\textwidth}
        \centering
        \textbf{Sujith Vemishetty} \\
        Synechron \\
        \texttt{sujith.vemishetty@synechron.com}
        \texttt{sujith.3920@gmail.com}
    \end{minipage}
    \begin{minipage}{0.5\textwidth}
        \centering
        \textbf{Advitiya Arora} \\
        Synechron \\
        \texttt{advitiya.arora@synechron.com}
        \texttt{adi23arora@gmail.com}
    \end{minipage}\\[3em]
    \begin{minipage}{0.5\textwidth}
        \centering
        \textbf{Anupama Sharma}\textsuperscript{\textdagger}\\
        \texttt{anupamasharma1112@gmail.com}
    \end{minipage}
}

\date{June 2025}

\vspace{0.5cm}

\begin{document}

\maketitle
\renewcommand{\thefootnote}{\textdagger}
\footnotetext{Work done while at Synechron.}
\thispagestyle{plain}
\pagestyle{plain}

\begin{abstract}
   The advancements in the domain of LLMs in recent years have surprised many, showcasing their remarkable capabilities and diverse applications. Their potential applications in various real-world scenarios have led to significant research on their reliability and effectiveness. On the other hand, multimodal LLMs and Text-to-Image models have only recently gained prominence, especially when compared to text-only LLMs. Their reliability remains constrained due to insufficient research on assessing their performance and robustness. This paper aims to establish a comprehensive evaluation framework for Text-to-Image models, concentrating particularly on their adherence to prompts. We created a novel dataset that aimed to assess the robustness of these models in generating images that conform to the specified factors of variation in the input text prompts. Our evaluation studies present findings on three variants of Stable Diffusion \cite{esser2024scaling} models: Stable Diffusion 3 Medium, Stable Diffusion 3.5 Large, and Stable Diffusion 3.5 Large Turbo, and two variants of Janus \cite{chen2025janus} models: Janus Pro 1B and Janus Pro 7B. We introduce a pipeline that leverages text descriptions generated by the \texttt{gpt-4o} model for our ground-truth images, which are then used to generate artificial images by passing these descriptions to the Text-to-Image models. We then pass these generated images again through \texttt{gpt-4o} using the same system prompt and compare the variation between the two descriptions. Our results reveal that these models struggle to create simple binary images with only two factors of variation: a simple geometric shape and its location. We also show, using pre-trained VAEs on our dataset, that they fail to generate images that follow our input dataset distribution.
\end{abstract}

\section{Introduction}

Text-to-Image models are increasingly becoming popular for their ability to synthesize high-fidelity images guided by complex natural language prompts. They are able to create stunning visuals, both realistic and fictional. Open models like Stable Diffusion \cite{esser2024scaling}, Playground \cite{liu2024playground}, Flux \cite{flux}, Janus Pro \cite{chen2025janus}, etc., have advanced the state-of-the-art in Text-to-Image generation models in terms of vividness, fidelity, and the complexity of images generated. In this paper, we evaluate the robustness in performance of a few selected state-of-the-art Text-to-Image models with respect to \textit{prompt adherence}.  Specifically, we evaluate their image generations in terms of robustness in adhering to the simple factors of variation provided in their text prompt for a single dataset.

The image generation pipeline in diffusion models is controlled by various parameters, one of which is the guidance scale. A higher guidance scale encourages models to generate images that are more closely linked to the given text prompt, making them adhere to the prompt as much as possible, albeit at the cost of curtailing their creativity. In this study, we explicitly used a very high guidance scale \textbf{(9.0)} for Stable Diffusion models to enforce strict prompt adherence. The creators of these models have claimed top-tier performance in prompt adherence (exceptional prompt adherence in just 4 steps for Stable Diffusion 3.5 Large Turbo) \cite{sd3.5_release}.

We have also evaluated Janus Pro models (1B and 7B) in this study. Janus Pro\cite{chen2025janus} is a unified multimodal model that decouples visual encoding for understanding and generation, employing an auto-regressive transformer architecture. This design enables Janus Pro to excel in both multimodal understanding and Text-to-Image instruction-following, achieving state-of-the-art performance in prompt adherence and compositional generation tasks while maintaining a unified, efficient framework. We do not have explicit control over the guidance scale or resolution parameters. Hence, we used the default settings for Janus models.

Hartwig et al. list a set of Text-to-Image evaluation metrics in \cite{hartwig2024evaluating}. They categorize image generation evaluation metrics into image-based and Text-Image alignment metrics. Most of these metrics focus on realism, aesthetics, and compositional quality. Image-based metrics assess image quality without considering textual input, while Text-Image alignment metrics examine how well a generated image aligns with its textual prompt. These can be either embedding-based or content-based. For example, CLIPScore \cite{hessel2021clipscore}, based on CLIP \cite{radford2021learning}, uses the cosine similarity between the text embedding vector and the image embedding vector (CLIP distance) to compute the alignment of text and image. Another metric, TIFA \cite{hu2023tifa}, is a content-based Text-Image matching metric. 

Before the introduction of CLIP \cite{radford2021learning}, Text-to-Image alignment was indirectly evaluated using Image-to-Text alignment models, which relied heavily on human-annotated datasets due to the limited capabilities of the then existing Image-to-Text models compared to current SOTA models such as \texttt{gpt-4o}. Early metrics such as SPICE \cite{anderson2016spice}, ViLBERTScore \cite{lee-etal-2020-vilbertscore}, and TIGEr \cite{jiang2019tiger} mainly compared generated captions with reference captions to assess alignment. Subsequently, direct content-based metrics emerged, allowing the decomposition of evaluation into aspects like object accuracy (OA), spatial relationships (S), nonspatial relationships (NS), and attribute bindings (AB). SeeTRUE (VNLI) \cite{yarom2024you} approached this problem by fine-tuning multimodal models like BLIP2 \cite{li2022blip} and PaLI17B \cite{chen2022pali} to compute alignment scores based on binary (yes / no) responses to Text-Image entailment prompts. Other metrics like SeeTRUE (VQ2) \cite{yarom2024you}, TIFA \cite{zhang2022perceptual}, VQAScore \cite{lin2024evaluating}, and DA-Score \cite{singh2023divide} employed Visual Question Answering (VQA) models to evaluate disjoint parts of prompts by generating and answering questions derived from the image prompts, enhancing detection of alignment in complex scenarios. \cite{lin2024evaluating} incorporates a fine-tuned VQA model to predict answer likelihoods, adopting a pre-trained FlanT5 \cite{chung2022scaling} and combining it with a pre-trained CLIP vision-encoder. Semantic Object Alignment (SOA) \cite{hinz2020semantic} and VISOR \cite{gokhale2022benchmarking} relied on pre-trained object detection models to validate object presence and attributes, with VISOR variants emphasizing more on spatial relationships. Text-Image Alignment Metric (TIAM) \cite{grimal2024tiam} employed a pre-trained segmentation model to characterize the contents of a set of generated images. Frameworks such as CompBench \cite{huang2023t2icompbench} and TISE \cite{dinh2022tise} provided composite metrics, combining multiple evaluation aspects.

These advancements offered more robust and nuanced evaluations, correlating closely with human judgments. However, reliance on pre-trained models posed challenges, including domain specificity and potential biases from overlapping training datasets between evaluation and generation models. Comprehensive frameworks like TISE\cite{dinh2022tise} addressed these limitations by consolidating multiple metrics into a unified evaluation package, achieving consistent rankings aligned with human evaluations.

We focus our evaluation only on prompt adherence to given factors of variation in the input. For this evaluation, we cannot use a dataset with a high-dimensional latent distribution for the following two reasons:

\begin{enumerate}
\item We want to utilize a trained VAE as a proxy to estimate the ability of Text-to-Image models to generate data that fits into a certain distribution when input prompts are descriptions of images belonging only to that distribution
\item We are generating the text descriptions of images using \texttt{gpt-4o}, which are fed into the Text-to-Image models as input prompts; hence, a dataset with a high-dimensional latent distribution can have high noise in these text descriptions
\end{enumerate}

Therefore, we created a new dataset, inspired by the dSprites dataset, which is widely used in measuring disentanglement in VAEs and GANs. The reasons for creating a new dataset and the details of the dataset are covered in Section \ref{sec:dataset}.

We use this dataset to train VAEs (see Section \ref{sec:appendix_vae} for details) and sample a test subset from it to evaluate the robustness of Text-to-Image models in generating simple binary images with well-defined geometric shapes at a given position when they are described by an image understanding model like \texttt{gpt-4o}.
Our contributions:
\begin{enumerate}
\item Create a new dataset inspired by dSprites that can be used to evaluate the robustness of Text-to-Image models with respect to their faithfulness in generating images adhering to the factors of variation included in the input text prompts
\item A novel evaluation methodology and pipeline combining both image understanding and image generation models to automate the process of evaluating the robustness of Text-to-Image models
\end{enumerate}

\section{Related Work}

\subsection{\textbf{Diffusion Models}}
Diffusion Models have become the most popular deep generative models for image synthesis in the past few years. Denoising Diffusion Probabilistic Models (DDPM) \cite{ho2020denoising}, introduced in 2020, have presented state-of-the-art performance in high-quality image synthesis. Subsequent research utilizing this technique for better Text-to-Image generation models has pushed the boundaries of imaginative image generation. The class of models called Stable Diffusion, created by StabilityAI, has been the most discussed in this domain in recent years, with their Stable Diffusion 3 class of models outperforming most others in human preference evaluations. These models utilize a technique called Rectified Flow, and their Multi-Modal Diffusion Transformer (MMDiT) architecture provides a better way to combine both image and text modalities so that the generated images can be closer to the input prompt, as they have more context from the text representation. The guidance scale parameter in these models controls the level of inventiveness, with a higher guidance scale restricting it and making the generated image follow the input prompt much more closely.

\subsection{\textbf{Unified Multi-Modal Models: Janus and Janus-Pro}}
Recent advances in unified multimodal models have demonstrated significant progress in both multimodal understanding (e.g., image captioning, visual question answering) and visual generation (e.g., Text-to-Image synthesis). Janus\cite{ma2024janusflow}, a pioneering model in this space, proposed a novel architecture that decouples visual encoding for multimodal understanding and generation tasks. This decoupling addresses a key limitation in prior approaches, where using a single visual encoder for both tasks led to suboptimal performance due to the differing representation requirements of understanding and generation. Janus was validated at a 1.5B parameter scale, but its effectiveness was limited by data scale and model size, particularly in the stability and quality of Text-to-Image generation.

Janus-Pro builds on this foundation by introducing three key improvements: (1) an optimized training strategy that refines the staging of adaptor, pre-training, and fine-tuning phases (2) a significant expansion of both multimodal understanding and visual generation datasets, including the addition of high-quality synthetic data for improved aesthetic image outputs (3) scaling the model up to 7B parameters, which improves convergence and performance across both understanding and generation tasks. The Janus-Pro architecture, similar to Janus, employs separate encoders and adaptors for understanding and generation, feeding their outputs into a unified autoregressive transformer for joint processing.

Empirical results show that Janus-Pro-7B achieves state-of-the-art performance on multimodal understanding benchmarks (e.g., MMBench, MME, GQA) and instruction-following visual generation benchmarks (e.g., GenEval, DPG-Bench), outperforming both unified and task-specific models such as Stable Diffusion 3 Medium, DALL-E 3, and LLaVA.

\subsection{\textbf{Variational Autoencoders}}
\textbf{VAEs} have been among the state-of-the-art techniques for generative image models, with significant research conducted over the years to improve their performance in disentanglement and reconstruction. The ${\beta}$-TCVAE \cite{chen2018isolating} technique, introduced in 2019, has shown state-of-the-art performance in disentanglement for established datasets like dSprites. The authors have found a novel way to isolate sources of disentanglement by decomposing the \textbf{Evidence Lower Bound} (\textbf{ELBO}) loss in VAEs and have shown that using a weighting parameter to control a term they call \textbf{Total Correlation Loss} (correlation between latent variables in ELBO) can achieve better disentanglement with proper hyperparameter tuning. We discuss the details of our VAE architecture and losses in Appendix A (Sec \ref{sec:appendix_vae}).

\subsection{\textbf{Latent Space Understanding Evaluation}}
To the best of our knowledge, there has not been any evaluation of the performance of Text-to-Image models with regard to their faithfulness toward the factors of variation given in the input text prompts. The closest related work to ours was done by Zhu et al. in \cite{latent_explainer} and \cite{explain_latent_representation}. Both works only deal with evaluating whether image-understanding Multimodal LLMs are able to understand and explain the latent variable that is changing in a given sequence of images, and they prove that \texttt{gpt-4o} is able to do a good job in explaining latent variable changes. Our work differs from these in the following ways:
\begin{enumerate}
\item Both \cite{latent_explainer} and \cite{explain_latent_representation} only evaluate the quality of the explanation generated by an image-understanding Large Multimodal Model (LMM), while we assume good quality of explanation generation by image-understanding LMMs and evaluate the prompt adherence capacity of Text-to-Image models
\item We also use the reconstruction loss of trained VAEs to evaluate the capability of Text-to-Image models to generate from a predefined latent variable distribution of a given dataset of images when the prompts are sampled from explanations of that same image dataset
\end{enumerate}

\subsection{\textbf{Text-to-Image and Image Understanding Evaluation}}
\begin{enumerate}
\item {\textbf{X-IQE: eXplainable Image Quality Evaluation}} \cite{chen2023x} employs pre-trained Vision-Language Large Models (VLLMs) to generate conversational analysis texts for images. The foundational model is MiniGPT-4, which is improved by a well-structured hierarchical Chain of Thought (CoT) scheme and strict format constraints that ensure outputs are correct and make sense. X-IQE evaluates image quality across three dimensions: fidelity, alignment, and aesthetics. MiniGPT-4 generates image descriptions, which are compared against initial captions or task-specific inputs. This approach ensures that the generated descriptions provide a robust assessment of the quality of the image by focusing on task-based attributes. By leveraging MiniGPT-4\textquotesingle s conversational capabilities and its structured reasoning, X-IQE delivers a comprehensive evaluation of image quality in diverse contexts.

\item {\textbf{MMGenBench}}'s work \cite{huang2024mmgenbench} seems close to our evaluation, but their pipeline caters to evaluating Large MultiModal models' image comprehension capabilities. They introduce a pipeline that leverages Text-to-Image models to generate new images based on LMM-generated prompts, aiming to evaluate LMM performance in comprehending and describing images within the generative image domain. To ensure robustness, the evaluation avoids the randomness inherent in Text-to-Image model inference by performing a representational-level comparison between generated and input images. Using the Unicom model for image embeddings, the study computes a similarity score (SIM-Score) to assess feature alignment and a generation quality score (FID) to evaluate the distribution similarity between generated and input images. This work is different from ours in its goal and the use of embedding-level comparison between the generated images and generated text embeddings. We learned about this work later in our experimentation, as the timing of the paper shows, and our study remains independent of it.

\item {\textbf{TIFA (Text-to-Image Faithfulness Evaluation with Question Answering)}} \cite{hu2023tifa} focuses on evaluating the faithfulness of generated images to their textual inputs through a Visual Question Answering (VQA) approach. GPT-3 is used to automatically generate question-answer pairs from the text input, which are then verified and filtered. A VQA system evaluates the generated image by answering these questions based on its content. The VQA system measures the faithfulness of the image to the text input by measuring the accuracy of its answers. TIFA shows a significantly higher correlation with human judgments compared to CLIPScore. However, TIFA performance is highly dependent on the capabilities of the underlying language and VQA models.
\end{enumerate} 

While recent advances in Text-to-Image models and multimodal LLMs have significantly improved the quality and faithfulness of image generation and understanding, critical gaps remain. Specifically, most evaluation frameworks focus either on the quality of generated images or on the explainability of latent representations, but do not systematically assess the alignment between generated images and the full spectrum of factors of variation specified in complex prompts. This motivates our approach, which integrates rigorous evaluation of prompt faithfulness, leveraging both VAE reconstruction and LMM explanations.

\section{Dataset}\label{sec:dataset}

We have created a new dataset for our evaluation experiments, inspired by the dSprites dataset proposed in \cite{dsprites_data}. The existing dSprites dataset had the following limitations for use in our experiments:

\begin{enumerate}
\item The images are of 64x64 resolution, making them obsolete for comparison with the high-resolution images currently generated by state-of-the-art Text-to-Image models (Refer Section \ref{sec:appendix_visualization_sample} )
\item The oval and heart shapes are not suitable for assessing the robustness of their generation based on simple text descriptions. This is because the heart shape has various definitions, and its orientation is difficult to describe verbally
\item The original dSprites dataset was generated in the Love2D framework, and its rasterization has induced noise in the shapes. Additionally, the original implementation is now unavailable, causing problems with reproducing a similar dataset at different resolutions
\end{enumerate}


\begin{figure*}[h!]
\centering
\begin{subfigure}[b]{0.3\textwidth}
\centering
\includegraphics[width=\textwidth]{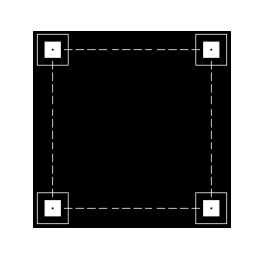}
\caption{Squares}
\label{fig:new_dSprites_sq_latents_viz}
\end{subfigure}%
~
\begin{subfigure}[b]{0.3\textwidth}
\centering
\includegraphics[width=\textwidth]{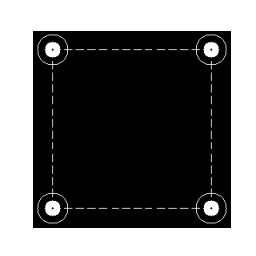}
\caption{Circles}
\label{fig:new_dSprites_cir_latents_viz}
\end{subfigure}%
~
\begin{subfigure}[b]{0.3\textwidth}
\centering
\includegraphics[width=\textwidth]{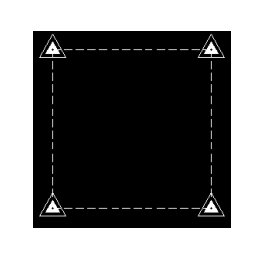}
\caption{Triangles}
\label{fig:new_dSprites_tri_latents_viz}
\end{subfigure}%
\caption{Visualization of shapes for the minimum and maximum latent values for Scale, Position X, and Position Y}
\label{fig:new_dSprites_latents_viz}

\end{figure*}

We remedy these limitations by creating a 256x256 resolution dataset that contains simple geometric shapes that are easily explained and generated. We generated a dataset with a total of 1,400,388 images. The latent values used for generating the images are visualized in figure \ref{fig:new_dSprites_latents_viz}. The differences in our dataset compared to dSprites are:

\begin{enumerate}
\item Our images are 256x256 resolution, while dSprites has 64x64 resolution, as many Text-to-Image models cannot generate any discernible information in a 64x64 resolution
\item We use simple, non-ambiguous geometric shapes in the dataset, viz. Square, Circle, and Triangle, instead of an ambiguously defined shape like Heart in dSprites
\item We use only a single orientation in the dataset (0 rotation) instead of 40 orientations in dSprites, because image understanding models are not capable of detecting the exact orientation value of a shape in the image
\item We created a dataset with 206 values for both Position X and Position Y. Using the same 1-pixel translation method as dSprites, we generated images of shapes centered from (0,0) to (205,205), leaving a 25-pixel margin on all sides
\end{enumerate}

\begin{figure*}[t!]
\centering
\begin{subfigure}[b]{0.3\textwidth}
\centering
\includegraphics[width=\textwidth]{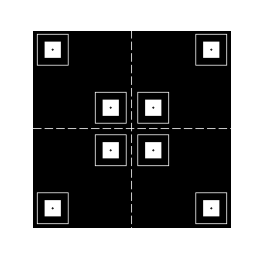}
\caption{}
\label{fig:new_dSprites_subset_sq_latents_viz}
\end{subfigure}%
~
\begin{subfigure}[b]{0.3\textwidth}
\centering
\includegraphics[width=\textwidth]{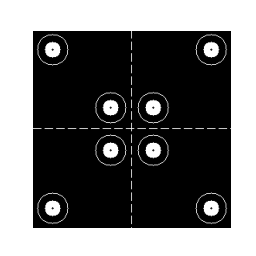}
\caption{}
\label{fig:new_dSprites_subset_cir_latents_viz}
\end{subfigure}%
~
\begin{subfigure}[b]{0.3\textwidth}
\centering
\includegraphics[width=\textwidth]{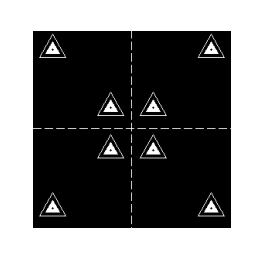}
\caption{}
\label{fig:new_dSprites_subset_tri_latents_viz}
\end{subfigure}%
~
\caption{Visualization of shapes in the test subset along with the center lines to show their closest and farthest positions from the center of the image in each quadrant}
\label{fig:new_dSprites_sample_data_latents_viz}
\end{figure*}

We train VAEs on the entire 1,400,388 images dataset, but we use only a sample of 9,216 images (3 shapes, 3 scales, 32 Position X, 32 Position Y) for evaluating the faithfulness criterion of Text-to-Image models towards the factors of variation. Specifically, we evaluate if the models are able to generate images with a specified shape in the prompt with the center of the shape falling in either of the 4 quadrants, viz. Top-Left, Top-Right, Bottom-Left, Bottom-Right. The latent values for this test sample dataset are visualized in Figure \ref{fig:new_dSprites_sample_data_latents_viz}.

We selected this sample of 9,216 images for evaluation due to the following constraints and reasons:

\begin{enumerate}
\item Experiments of this nature on much larger datasets will take up too many resources due to the compute-intensive nature of image understanding LMMs and Text-to-Image models
\item We sampled 3 scales out of 11 in the original dataset, as it is a reasonable assumption that if the performance is evaluated and averaged for the smallest, middle, and largest scales, it will be representative of all the scales
\item The current image understanding LMMs can only generate a description of the image with an approximate position of the shape categorized into 4 quadrants of the image, and they do not or cannot give a description depicting the exact pixel value position of the center of shape
\item We leave a margin ($\approx$50px) with respect to the center of the image in both the horizontal and vertical directions when sampling the X and Y translations of shapes, because the image understanding model cannot properly describe the quadrant of the center point of the shape when it is too close to the middle in either the horizontal or vertical direction. See Figure \ref{fig:new_dSprites_sample_data_latents_viz} for further clarity.
\end{enumerate}
\section{Methodology}
Our evaluation methodology begins with our sample dataset of 9,216 images as described in Section \ref{sec:dataset}. We generate text descriptions for these images using \texttt{gpt-4o}. Additionally, we predict the shape present in each image and identify the quadrant where the center of that shape is located. The prompt used for generating these descriptions is provided in Section \ref{sec:appendix_configuration}.

The overall methodology to generate our evaluation data can be described as follows:

\begin{enumerate}
    \item Text Description Generation: \texttt{gpt-4o} is used to create text descriptions for the ground truth images, including predictions of the shapes present and their respective quadrants
    
    \item Image Generation: Images are generated based on the text descriptions from \texttt{gpt-4o} using the following models:
    \begin{itemize}
        \item Stable Diffusion 3 Medium
        \item Stable Diffusion 3.5 Large
        \item Stable Diffusion 3.5 Large Turbo
        \item Janus Pro 7B
        \item Janus Pro 1B
    \end{itemize}
    Using Stable Diffusion models images are produced at three resolutions: 256x256, 512x512, and 1024x1024. In addition, Janus Pro 7B and Janus Pro 1B are used to generate images at a resolution of 384x384

    \item Repeat Text Description Generation: For the images generated in \textit{step 2}, the process of generating text descriptions using \texttt{gpt-4o} is repeated

    \item Repeat Image Generation: For the descriptions obtained in \textit{step 3}, the corresponding images for each resolution and model are generated

    \item Final Text Description Generation: For the images produced in \textit{step 4}, process of generating text descriptions using \texttt{gpt-4o} is repeated
\end{enumerate}
The evaluation methodology involves measuring three key metrics based on the generated data.
\begin{enumerate}\label{evaluation_criteria}
    \item Reconstruction Loss: The increase in reconstruction loss across each iteration of image generation using trained VAEs. Note that images with resolutions of 512x512, 1024x1024, and 384x384 are resized to 256x256, as our VAEs are trained on a dataset of 256x256 images
    
    \item F1-Score Degradation (Shape Prediction): The degradation of the F1-Score for shape predictions made by \texttt{gpt-4o}
    
    \item F1-Score Degradation (Quadrant Prediction): The degradation of the F1-Score for quadrant predictions made by \texttt{gpt-4o}
\end{enumerate}

\section{Results}
We now present the results of our experiments for the test dataset according to the evaluation criteria mentioned in \ref{evaluation_criteria}. 
Figure \ref{fig:final_shape_f1} presents the F1-Scores reflecting the accuracy of \texttt{gpt-4o's} predictions about the shapes and their position in the images. We can observe from the ground truth value of 1.0 for shapes and 0.95 for quadrant (in which the shape is centered) that \texttt{gpt-4o} was able to predict these attributes in all the ground truth images with ~100\% accuracy.

The following sections cover in detail the inferences obtained from our evaluations of Stable Diffusion and Janus Pro models.  
Multiple factors demonstrate a direct correlation with our primary evaluation metrics: the F1 score for image attribute prediction and the reconstruction loss between original and generated images in our trained variational autoencoders (VAEs).
These factors comprise image resolution, iteration number and binarization threshold. We alias these factors as control variables. 

For both sections we first discuss the degradation of prediction quality in the textual descriptions generated by the \texttt{gpt-4o} model. Then we turn our attention to the degradation of reconstruction using pre-trained VAEs with the generated images for those descriptions.
\subsection{\textbf{Stable Diffusion Results}}
\begin{figure}[H]
    \centering
    \includegraphics[width=1\textwidth]{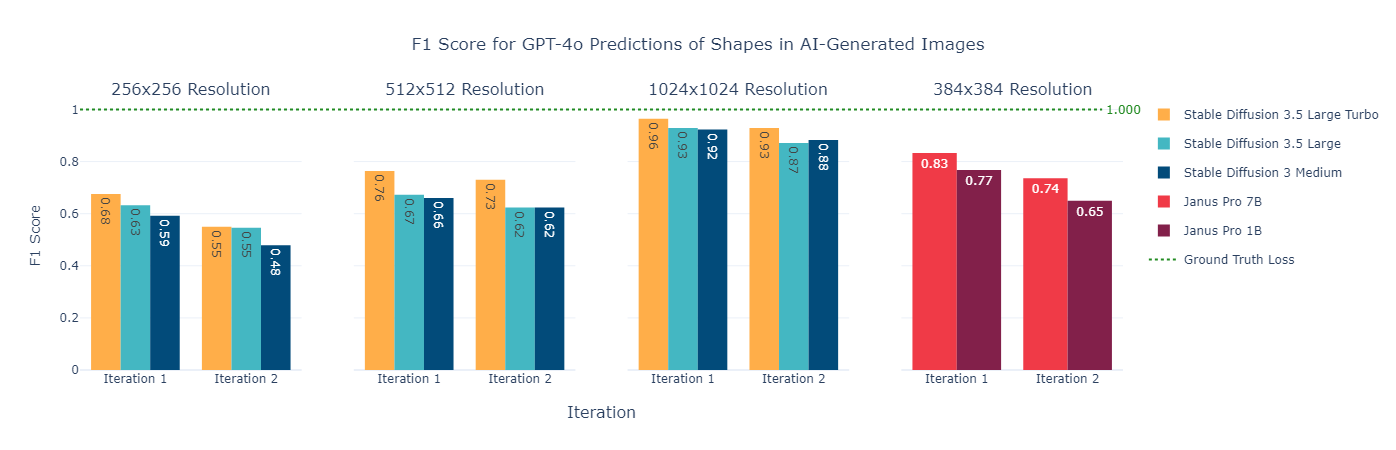} 
    \caption{}
    \label{fig:final_shape_f1} 
\end{figure}

\begin{figure}[H]
    \centering
    \includegraphics[width=1\textwidth]{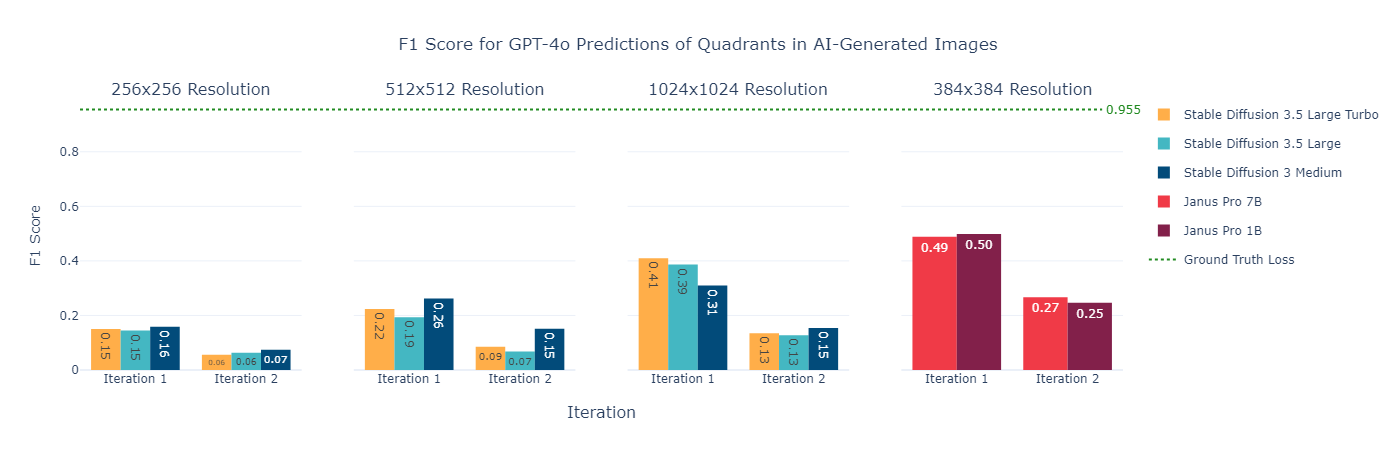}

    \caption{}
    \label{fig:final_quads_f1} 
\end{figure}

Here we shall compare the results across Stable Diffusion model variants and showcase the effects of the control variables. 
\begin{enumerate}
    \item {\textbf{Impact of Image Resolution on F1 score for Shape prediction}}:
    \begin{enumerate}
        \item Figure \ref{fig:final_shape_f1} shows that the highest F1-Score of 96\% was achieved in the first iteration for 1024x1024 resolution images, with all models reaching around 90\% accuracy in F1-Scores across both iterations. 
        \item For 512x512 resolution image generation, Stable Diffusion 3 Medium and Stable Diffusion 3.5 Large show similar performance. Stable Diffusion 3.5 Large Turbo demonstrates significantly better accuracy, with almost 10\% higher performance in both iterations. However, the highest F1-Score for 512x512 resolution is only 76\% for Stable Diffusion 3.5 Large Turbo in the first iteration, indicating a substantial decrease in prompt adherence for simple shape generation when moving from 1024x1024 to 512x512 resolution.
        \item Our ground truth images are in 256x256 resolution. However, while observing the results in 256x256 resolution, we infer that no model reached even 70\% F1-Score in the first iteration and remained $\approx$50\% in the second iteration. We can thus observe that the prompt adherence is minimal for even the simplest geometric shapes with straightforward binary images.
    \end{enumerate}
    Thus, we observe that prompt adherence drops significantly as we move to lower resolutions.
    \item {\textbf{Impact of Image Resolution on F1 Score for Quadrant prediction}}:
    \begin{enumerate}
        \item Figure \ref{fig:final_quads_f1} shows that compared to shape prediction, quadrant prediction proved to be substantially more challenging. The highest F1 score was only 0.41 and was achieved by Stable Diffusion 3.5 Large Turbo in the first iteration for 1024x1024 resolution. Stable Diffusion 3.5 Large and Stable Diffusion 3 Medium showed an even worse performance.
        \item At 512x512 resolution, Stable Diffusion 3 Medium surprisingly performed the best with an F1 score of 26\%. However, a considerable drop is seen as compared to 1024x1024 resolution
        \item At 256x256 resolution, the performance dropped even further.
    \end{enumerate}
    We notice a similar pattern as compared to shape prediction and observe a drop in prompt adherence as we move to lower resolutions.
    \item {\textbf{Impact of Iteration on F1 Score}}
    \begin{enumerate}
        \item Figure \ref{fig:final_shape_f1} also shows that keeping the resolution fixed, we observe that on average, the F1 score for shape prediction decreases (between 3\%-13\%)  as we move to further iterations. 
        \item Figure \ref{fig:final_quads_f1} shows that keeping the resolution fixed, we observe a drastic drop in F1 score (between 9\%-28\%) for quadrant prediction. The performance degrades as we move to further iterations.
    \end{enumerate}
    \item {\textbf{Impact of Binarization Threshold over Iterations on Reconstruction Loss}}
    \begin{enumerate}
        \item ${\beta}$-TCVAE proved to be the best VAE in terms of having the lowest reconstruction loss. Thus, we have shown the results for ${\beta}$-TCVAE in this section. Refer \ref{sec:Impact of Threshold over Iterations} to see the results for other VAEs.
        \item Keeping in mind that some aberrations exist, Figure \ref{fig:final_thresholding.png} shows that on average, binarization threshold doesn't affect the reconstruction loss much.
    \end{enumerate}
    \item {\textbf{Impact of Resolution on Reconstruction Loss}}
    \begin{enumerate}
        \item For images generated by Stable Diffusion models in 512x512 and 1024x1024 resolutions, the reconstruction loss is measured by down-scaling them to 256x256 resolution.
        \item Figure \ref{fig:final_model_iter_vs_avg_recon_loss.png} leads us to conclude that using high resolution images, in general, leads to a better reconstruction. 
        \item Refer to \ref{sec:Impact of Resolution over Iterations} to see results across other VAEs
    \end{enumerate}
    \item {\textbf{Impact of Iteration on Reconstruction Loss}}
    \begin{enumerate}
        \item Figure \ref{fig:final_model_iter_vs_avg_recon_loss.png} also shows the significant decline in reconstruction quality over subsequent iterations.
        \item Reconstruction quality for ground truth images is good as expected. This helps to verify that no errors have taken place in training the VAEs.
        \item However, as we progress in the loop of generating a text description of an image and using it to generate an artificial image, we clearly observe a loss of information in the text-to-image stage and lack of faithfulness towards the factors of variation in input text prompts, all of which leads to a worse reconstruction.
    \end{enumerate}
    We would like to emphasize that reconstruction loss values were quite similar across inferences on 6 out of 7 different models trained with MSE loss.
    This suggests that in terms of reconstruction quality, we have reached a saturation point. This is encouraging as it suggests that our VAEs have been trained adequately. Refer to Section \ref{sec:qualitative_evaluation} for a qualitative evaluation of our VAEs.
\end{enumerate}
\subsection{\textbf{Janus Pro Results}}
\begin{figure}[H]
    \centering
    \includegraphics[width=1\textwidth]{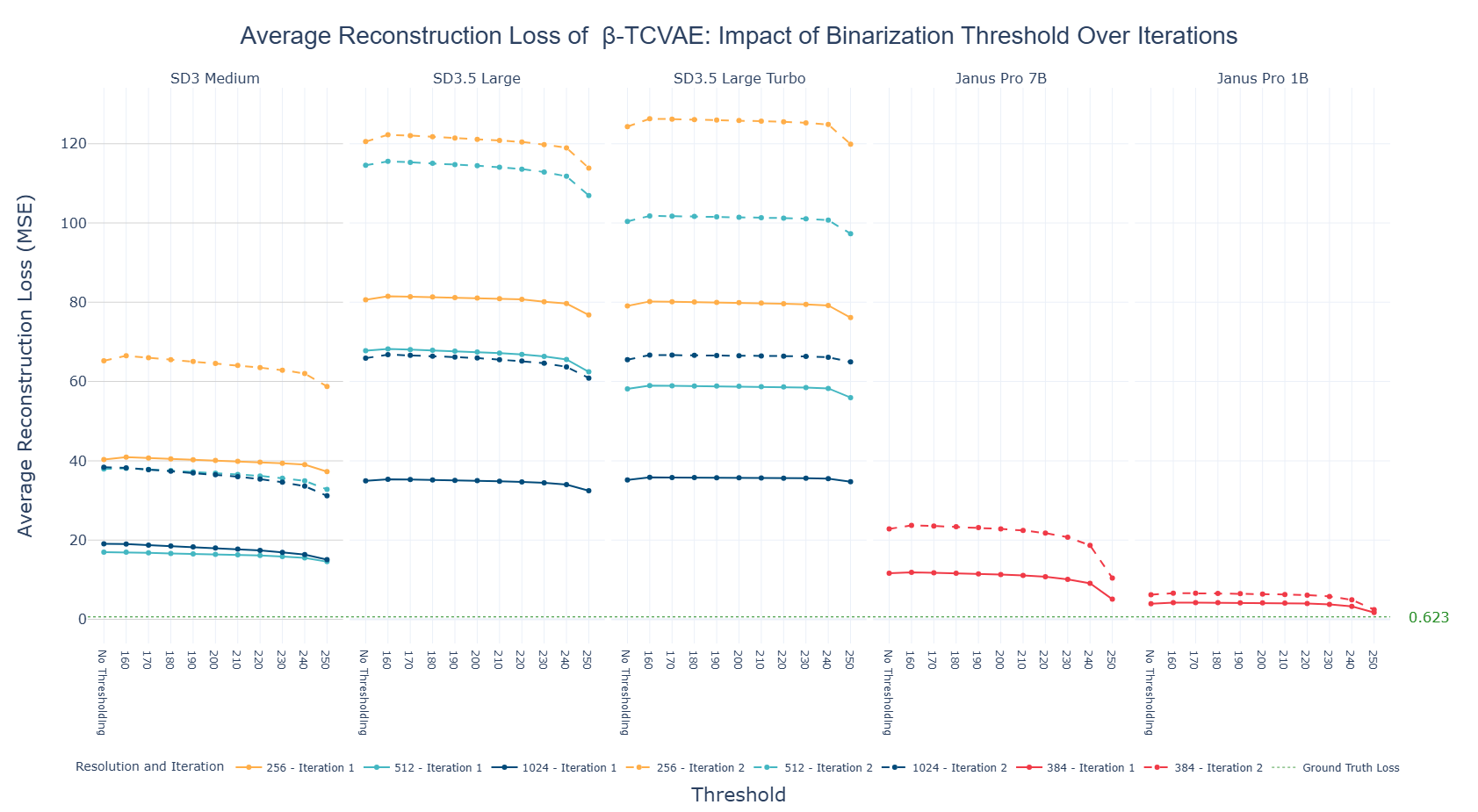}
    \caption{}
    \label{fig:final_thresholding.png} 
\end{figure}

\begin{figure}[H]
    \centering
    \includegraphics[width=1\textwidth]{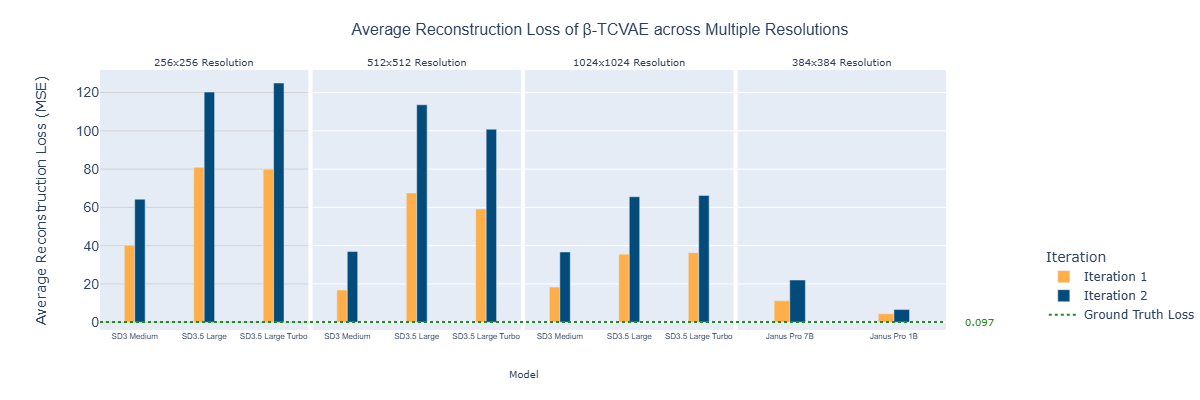}
    \caption{}
    \label{fig:final_model_iter_vs_avg_recon_loss.png} 
\end{figure}

Here we shall compare the results across Janus Pro model variants and showcase the effects of the control variables. Since Janus Pro only supports 384x384 resolution for image generation, any scope of comparing results across image resolution is eliminated.
\begin{enumerate}
    \item {\textbf{Impact of Iteration on F1 Score}}
    \begin{enumerate}
        \item Figure \ref{fig:final_shape_f1} shows that for shape prediction the F1 score decreases by almost ~10\% as we move to further iterations. As expected, Janus Pro 7B performed better than Janus Pro 1B in both iterations by a margin of 6$\%$ and 9$\%$ for iteration 1 and 2 respectively.
        \item Figure \ref{fig:final_quads_f1} shows that for quadrant prediction the F1 score drastically decreases by almost ~25\% as we move to further iterations. Surprisingly, both Janus Pro variants displayed a similar performance for both iterations. 
    \end{enumerate}
    \item {\textbf{Impact of Binarization Threshold over Iterations on Reconstruction Loss}}
    \begin{enumerate}
        \item With reference to ${\beta}$-TCVAE's reconstruction, Figure \ref{fig:final_thresholding.png} shows that on average, the binarization threshold does not affect the reconstruction loss much.
    \end{enumerate}
    \item {\textbf{Impact of Iteration on Reconstruction Loss}}
    \begin{enumerate}
        \item Figure \ref{fig:final_model_iter_vs_avg_recon_loss.png} also shows the significant decline in reconstruction quality over subsequent iterations for both Janus Pro variants.
        \item Similar to the Stable Diffusion models case, we clearly observe a loss of information in the text-to-image stage and lack of faithfulness towards the factors of variation in input text prompts leading to a worse reconstruction
    \end{enumerate}
\end{enumerate}
Evidence suggests that Janus Pro models do much better than Stable Diffusion variants despite being limited to lower resolution. Images generated by Janus Pro models are more faithful to prompts and lead to better reconstruction in multiple diverse scenarios. But we cannot state for sure the capability of Janus Pro models at higher resolutions until that feature is released. Looking at it more positively, both variants, if not limited by resolution show promising prompt adherence for generating images with a given shape. On the other hand, Figure \ref{fig:final_quads_f1} shows that the performance of all diffusion models and Janus pro models tested for generation of images with the shape lying in a given quadrant is completely inadequate for all resolution levels. The best F1-Score achieved for generating images with a shape in a given quadrant, using the same \texttt{gpt-4o} model, is only 0.41 for diffusion models and 0.5 for Janus Pro models, showcasing how these models struggle at understanding where the shape needs to be put in an image despite being provided with a high guidance scale configuration of 9.0. There are a few more interesting details which need to be pointed out.
\begin{enumerate}
    \item Stable Diffusion 3 Medium showed lower reconstruction loss than other model variants when tested with VAEs, and this pattern was consistent across all VAE architectures. However, closer examination revealed that this better performance was misleading because Stable Diffusion 3 Medium frequently generated blank images. Other model variants actually tried to recreate the target shapes, though they tended to produce images with more white pixels than the ground truth images, which contained smaller shapes. Since reconstruction loss is calculated using mean squared error (MSE) between the ground truth and generated pixels, the blank images from SD3 Medium likely created a bias that artificially lowered the loss scores. Because VAE reconstruction was only used as a supporting tool in our method for testing prompt adherence, we recommend not interpreting these reconstruction loss results as evidence of better image quality. Alternative evaluation metrics should be used for comparing these models.
    \item While predicting quadrants, we can observe that lighter models beat the heavier models. We would need further investigation to see if these patterns of one model outperforming the other at different resolution and iteration have any concrete reasoning behind or they are just chance based for this particular dataset. But the results do show that while the images generated by the three Stable Diffusion models ( at least in 1024x1024 resolution ) and Janus models can match the shape given in the text description well but no model has enough relative spatial awareness to generate the images with shapes positioned at a given location in an image. 
    \item Janus Pro models beat Stable Diffusion 3 variants in terms of prompt adherence and reconstruction loss tests. The actual reason for this requires further investigation into the architectures of the models.
\end{enumerate}
\section{Conclusion and Future Scope}
The Text-to-Image models that we tested are unable to generate images with rigid prompt adherence when it comes to simple binary images with only one geometric shape in them. The text prompts are straightforward and are being generated by LLMs which are also tokenization-based methods. We would need to further investigate how we can improve this performance. One solution can be a simple fine-tuning but it is not scalable for the infinitely many concepts present in the wild and is impractical for a general purpose Text-to-Image model. We would have to do a deeper analysis of the internal architecture of the Text-to-Image models to see which areas are impacting these aspects of prompt adherence the most and have to make the needed architectural modifications in the pre-training phase to have a scalable approach for the Text-to-Image models to gain conceptual understanding better.
\section{Acknowledgments}
We would like to extend our sincere thanks to our company Synechron for providing us the opportunity and necessary resources for this research. Special thanks to Bangalore Innovation team, Hareesha Pattaje and also Anantha Sharma for their support in this endeavour.
\newpage
\printbibliography 
\section{Appendix A: Sample generated images}\label{sec:appendix_visualization_sample}

In this section we show images generated at 4 different resolutions viz. 64x64, 256x256, 512x512, 1024x1024 for three sample prompts by all three stable diffusion models considered in our study. It can be noted that the generation capacity of the models at 64x64 resolution is severely limited in most cases. We have empirically verified that the details in images generated at 64x64 resolution are highly obscure and most of the times it can look like just random noise (although some models do generate some details at every resolution), we have shown only a single instance of such generation here. This observation has made us create a new dataset at 256x256 resolution instead of 64x64 so that all the models can be evaluated properly. We have limited ourselves to creating only 256x256 resolution dataset and not higher because the size of dataset will be too massive to handle with our available resources. 
\subsection{Prompt: Square}
Prompt: A black background with a small white square located in the top left corner. The square is solid filled and positioned near the edge of the image.

\begin{figure}[H]
    \centering
    \begin{tabular}{@{}ccc@{}}
        \textbf{Stable Diffusion 3 Medium} & \textbf{Stable Diffusion 3.5 Large} & \textbf{Stable Diffusion 3.5 Large Turbo} \\
        \fbox{\includegraphics[width=0.28\textwidth]{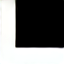}} &
        \fbox{\includegraphics[width=0.28\textwidth]{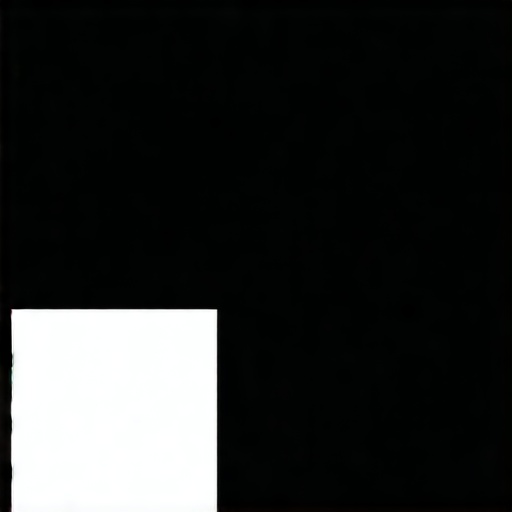}} &
        \fbox{\includegraphics[width=0.28\textwidth]{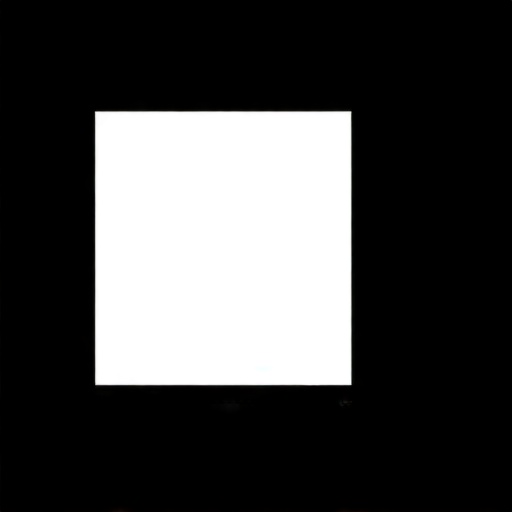}}
    \end{tabular}
    \caption{Sample images for prompt "square" at 64x64 resolution}
    \label{fig:square_res_64x64}
\end{figure}

\begin{figure}[H]
    \centering
    \begin{tabular}{@{}ccc@{}}
        \textbf{Stable Diffusion 3 Medium} & \textbf{Stable Diffusion 3.5 Large} & \textbf{Stable Diffusion 3.5 Large Turbo} \\
        \fbox{\includegraphics[width=0.28\textwidth]{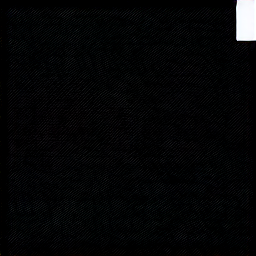}} &
        \fbox{\includegraphics[width=0.28\textwidth]{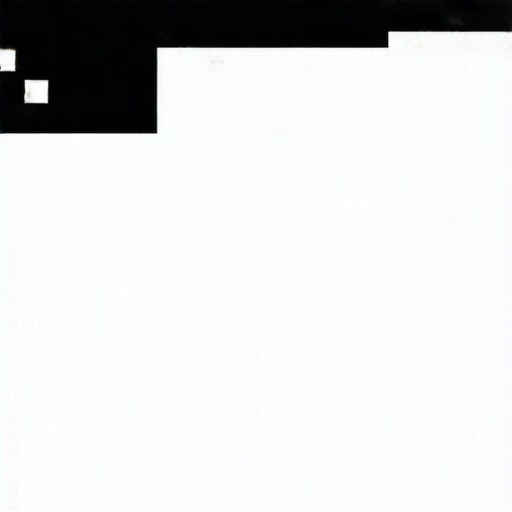}} &
        \fbox{\includegraphics[width=0.28\textwidth]{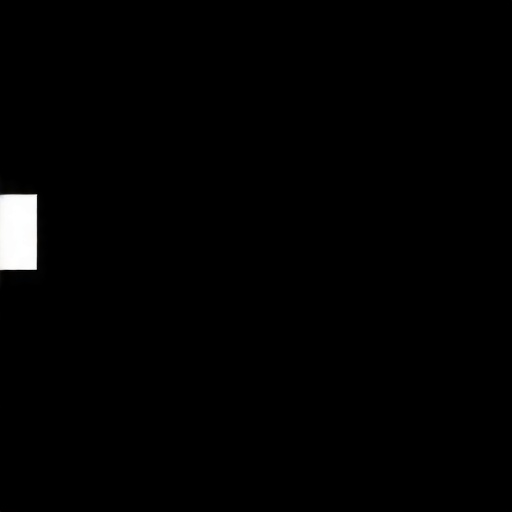}}
    \end{tabular}
    \caption{Sample images for prompt "square" at 256x256 resolution}
    \label{fig:square_res_256x256}
\end{figure}

\begin{figure}[htbp]
    \centering
    \begin{tabular}{@{}ccc@{}}
        \textbf{Stable Diffusion 3 Medium} & \textbf{Stable Diffusion 3.5 Large} & \textbf{Stable Diffusion 3.5 Large Turbo} \\
        \fbox{\includegraphics[width=0.28\textwidth]{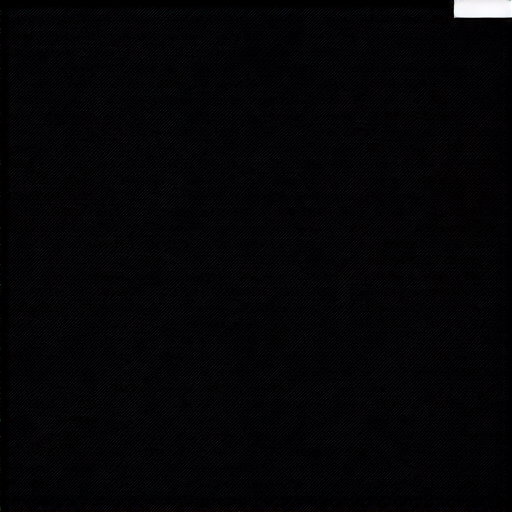}} &
        \fbox{\includegraphics[width=0.28\textwidth]{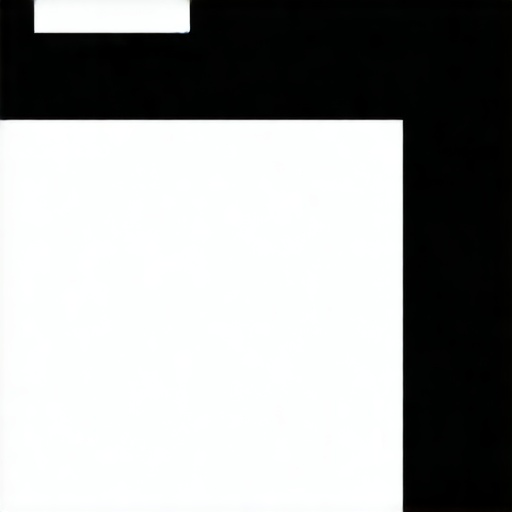}} &
        \fbox{\includegraphics[width=0.28\textwidth]{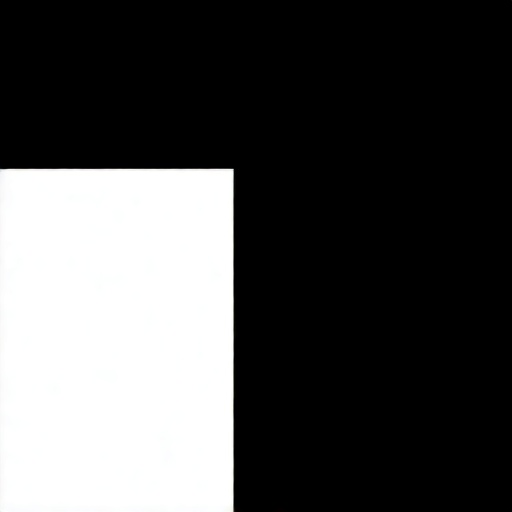}}
    \end{tabular}
    \caption{Sample images for prompt "square" at 512x512 resolution}
    \label{fig:square_res_512x512}
\end{figure}

\begin{figure}[htbp]
    \centering
    \begin{tabular}{@{}ccc@{}}
        \textbf{Stable Diffusion 3 Medium} & \textbf{Stable Diffusion 3.5 Large} & \textbf{Stable Diffusion 3.5 Large Turbo} \\
        \fbox{\includegraphics[width=0.28\textwidth]{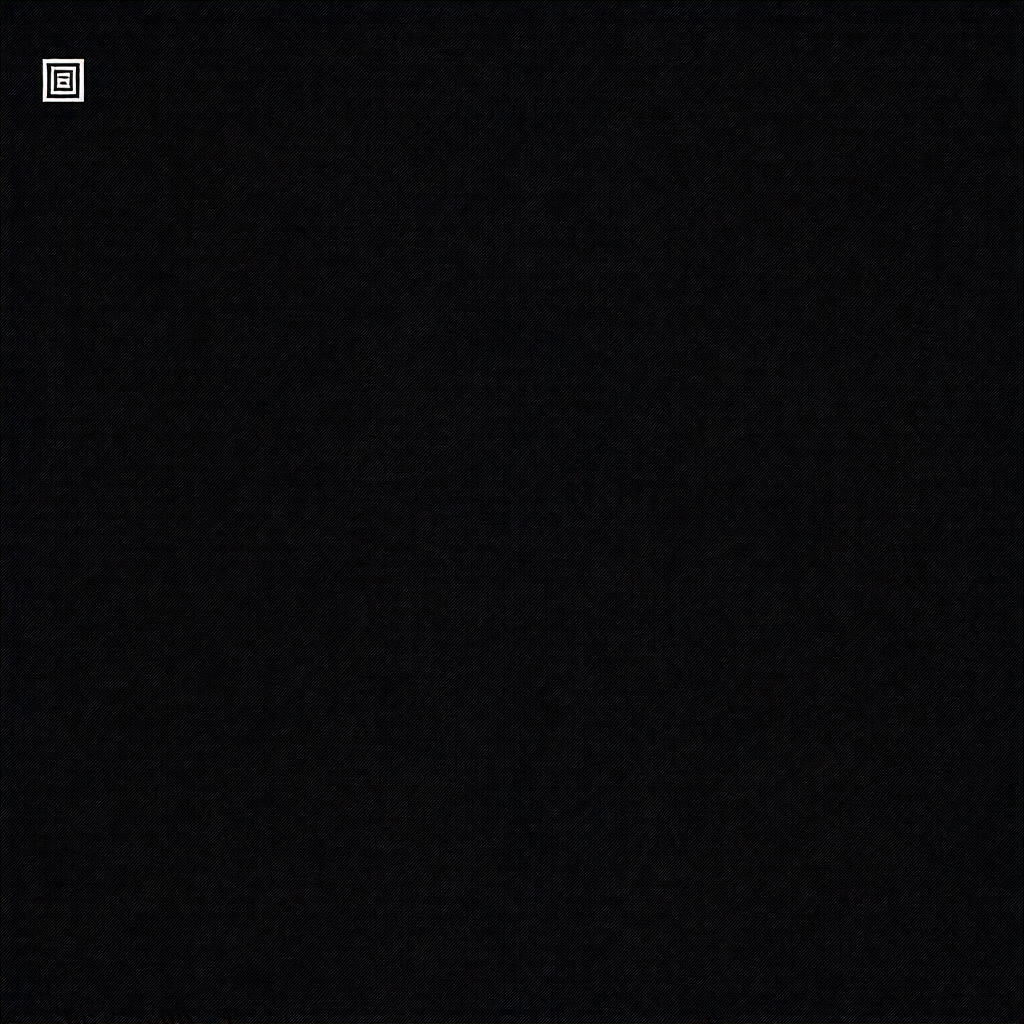}} &
        \fbox{\includegraphics[width=0.28\textwidth]{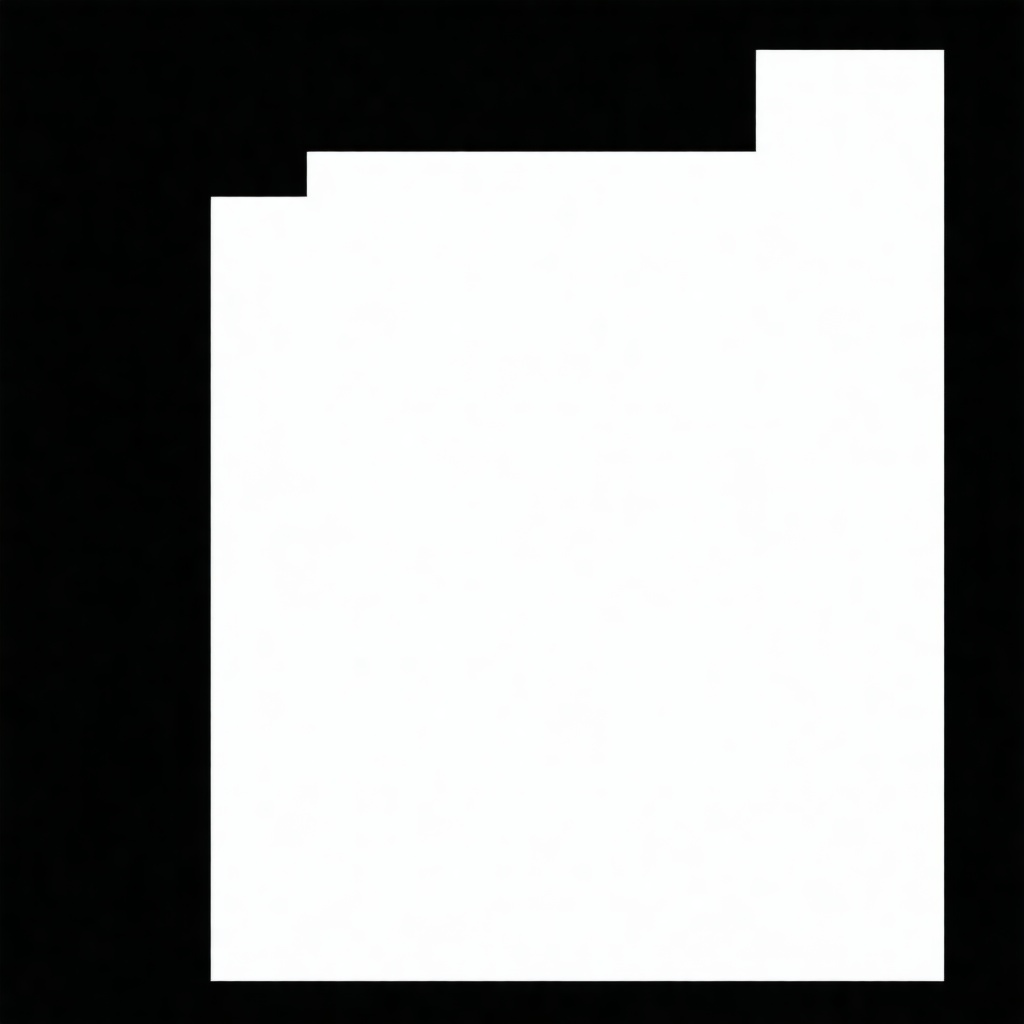}} &
        \fbox{\includegraphics[width=0.28\textwidth]{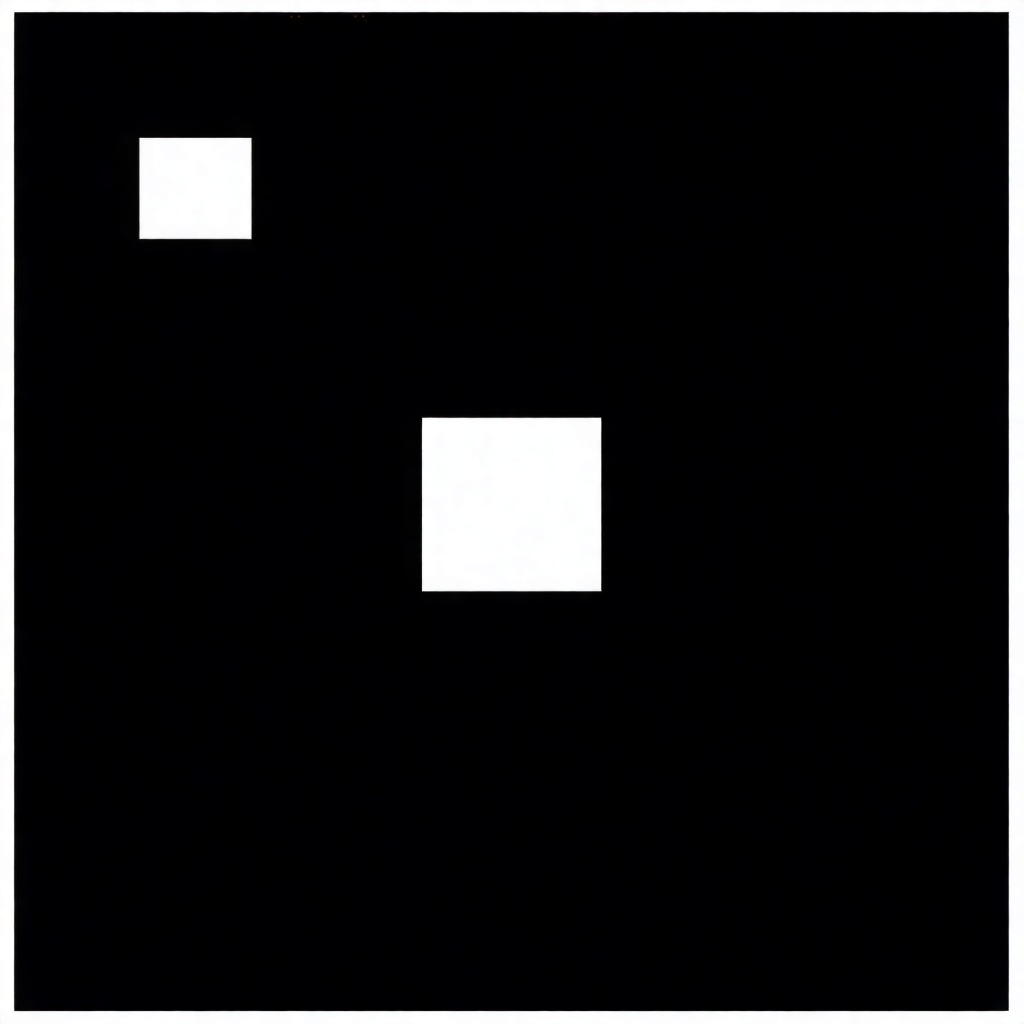}}
    \end{tabular}
    \caption{Sample images for prompt "square" at 1024x1024 resolution}
    \label{fig:square_res_1024x1024}
\end{figure}

\subsection{Prompt: Circle}
Prompt: A black background with a solid white circle located in the top left corner. The circle is small and fully visible, positioned near the edge of the image.

\begin{figure}[H]
    \centering
    \begin{tabular}{@{}ccc@{}}
        \textbf{Stable Diffusion 3 Medium} & \textbf{Stable Diffusion 3.5 Large} & \textbf{Stable Diffusion 3.5 Large Turbo} \\
        \fbox{\includegraphics[width=0.28\textwidth]{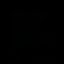}} &
        \fbox{\includegraphics[width=0.28\textwidth]{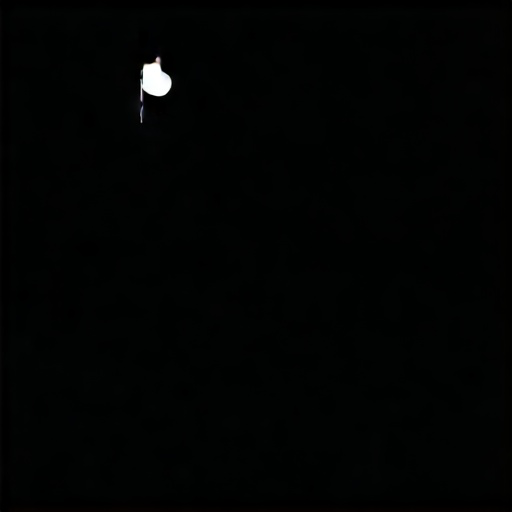}} &
        \fbox{\includegraphics[width=0.28\textwidth]{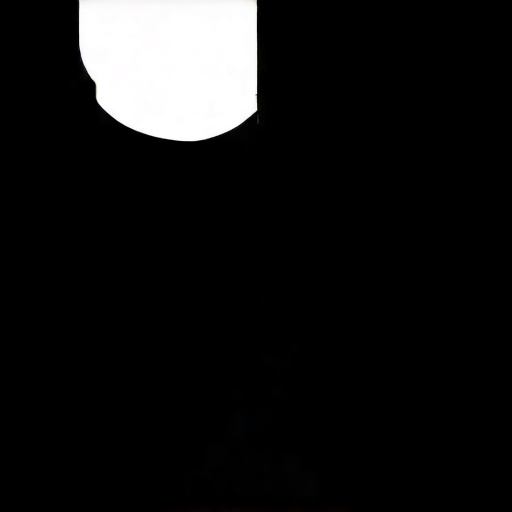}}
    \end{tabular}
    \caption{Sample images for prompt "circle" at 64x64 resolution}
    \label{fig:circle_res_64x64}
\end{figure}

\begin{figure}[htbp]
    \centering
    \begin{tabular}{@{}ccc@{}}
        \textbf{Stable Diffusion 3 Medium} & \textbf{Stable Diffusion 3.5 Large} & \textbf{Stable Diffusion 3.5 Large Turbo} \\
        \fbox{\includegraphics[width=0.28\textwidth]{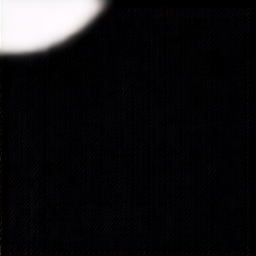}} &
        \fbox{\includegraphics[width=0.28\textwidth]{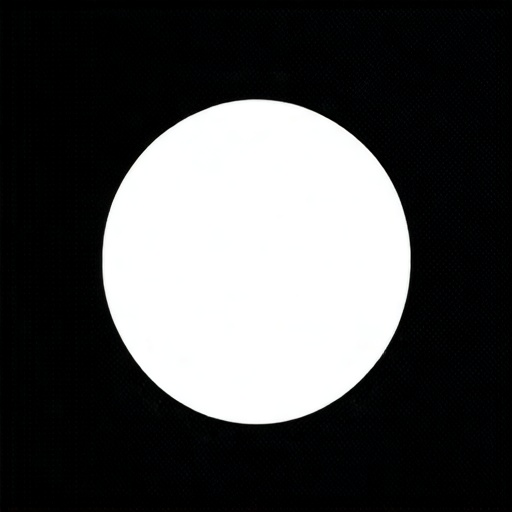}} &
        \fbox{\includegraphics[width=0.28\textwidth]{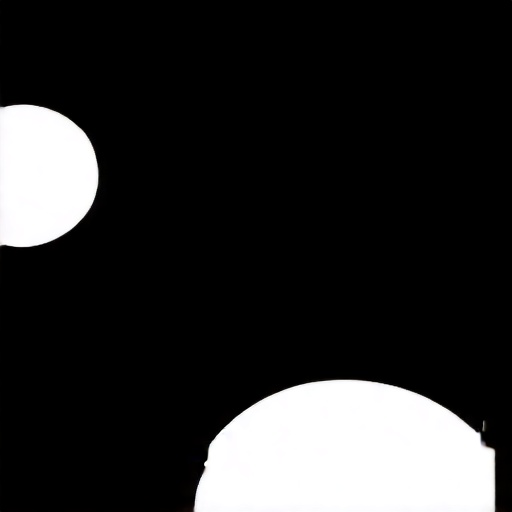}}
    \end{tabular}
    \caption{Sample images for prompt "circle" at 256x256 resolution}
    \label{fig:circle_res_256x256}
\end{figure}

\begin{figure}[htbp]
    \centering
    \begin{tabular}{@{}ccc@{}}
        \textbf{Stable Diffusion 3 Medium} & \textbf{Stable Diffusion 3.5 Large} & \textbf{Stable Diffusion 3.5 Large Turbo} \\
        \fbox{\includegraphics[width=0.28\textwidth]{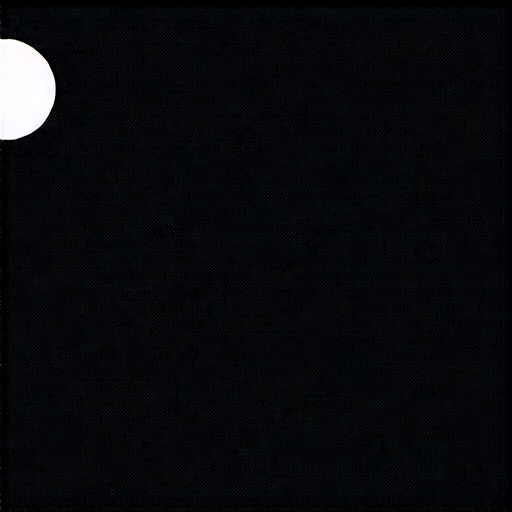}} &
        \fbox{\includegraphics[width=0.28\textwidth]{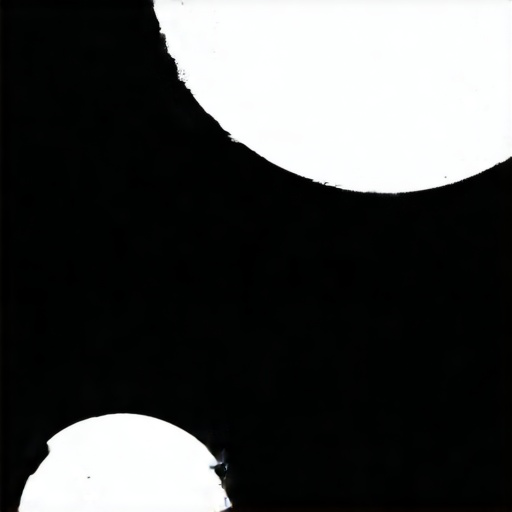}} &
        \fbox{\includegraphics[width=0.28\textwidth]{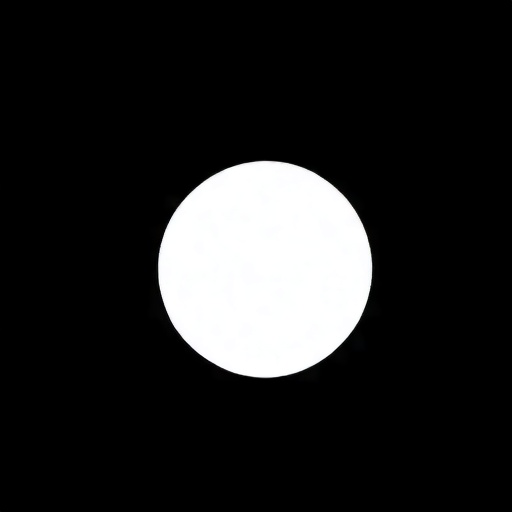}}
    \end{tabular}
    \caption{Sample images for prompt "circle" at 512x512 resolution}
    \label{fig:circle_res_512x512}
\end{figure}

\begin{figure}[H]
    \centering
    \begin{tabular}{@{}ccc@{}}
        \textbf{Stable Diffusion 3 Medium} & \textbf{Stable Diffusion 3.5 Large} & \textbf{Stable Diffusion 3.5 Large Turbo} \\
        \fbox{\includegraphics[width=0.28\textwidth]{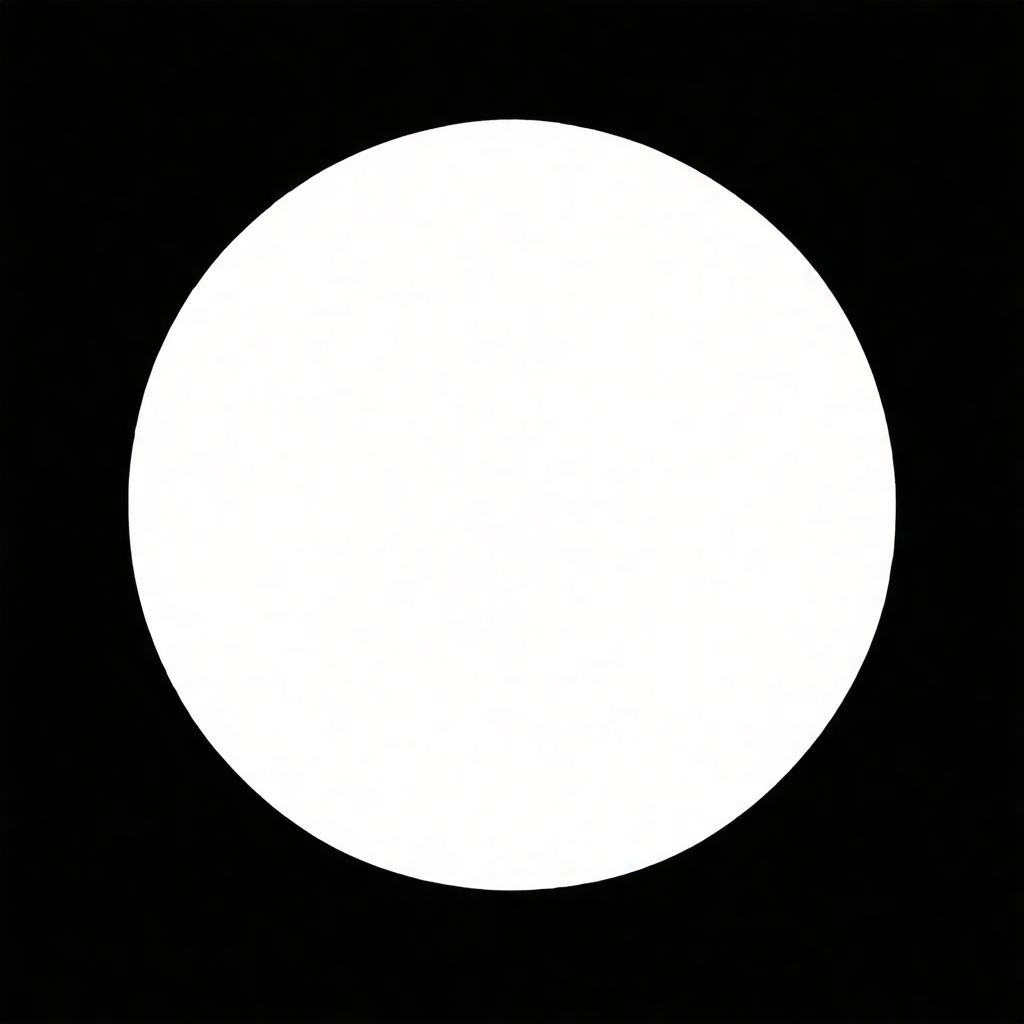}} &
        \fbox{\includegraphics[width=0.28\textwidth]{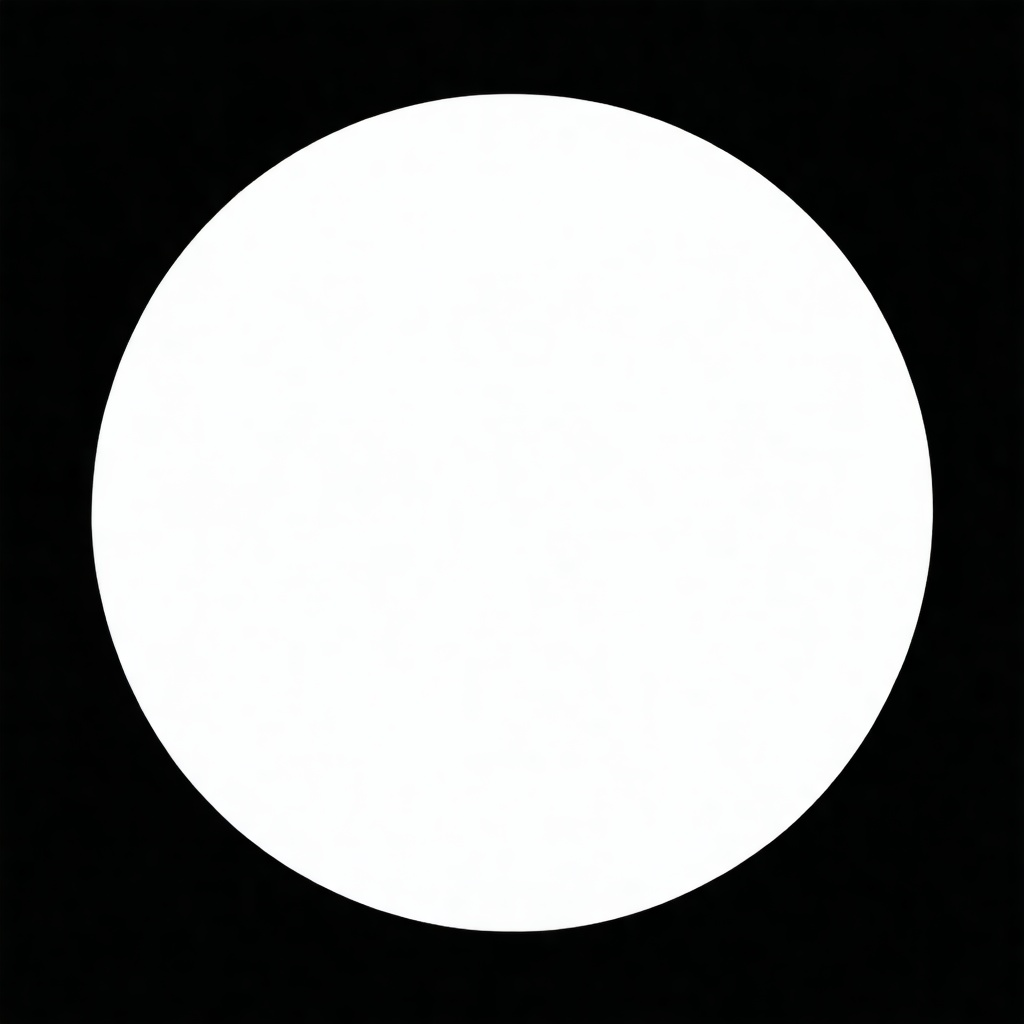}} &
        \fbox{\includegraphics[width=0.28\textwidth]{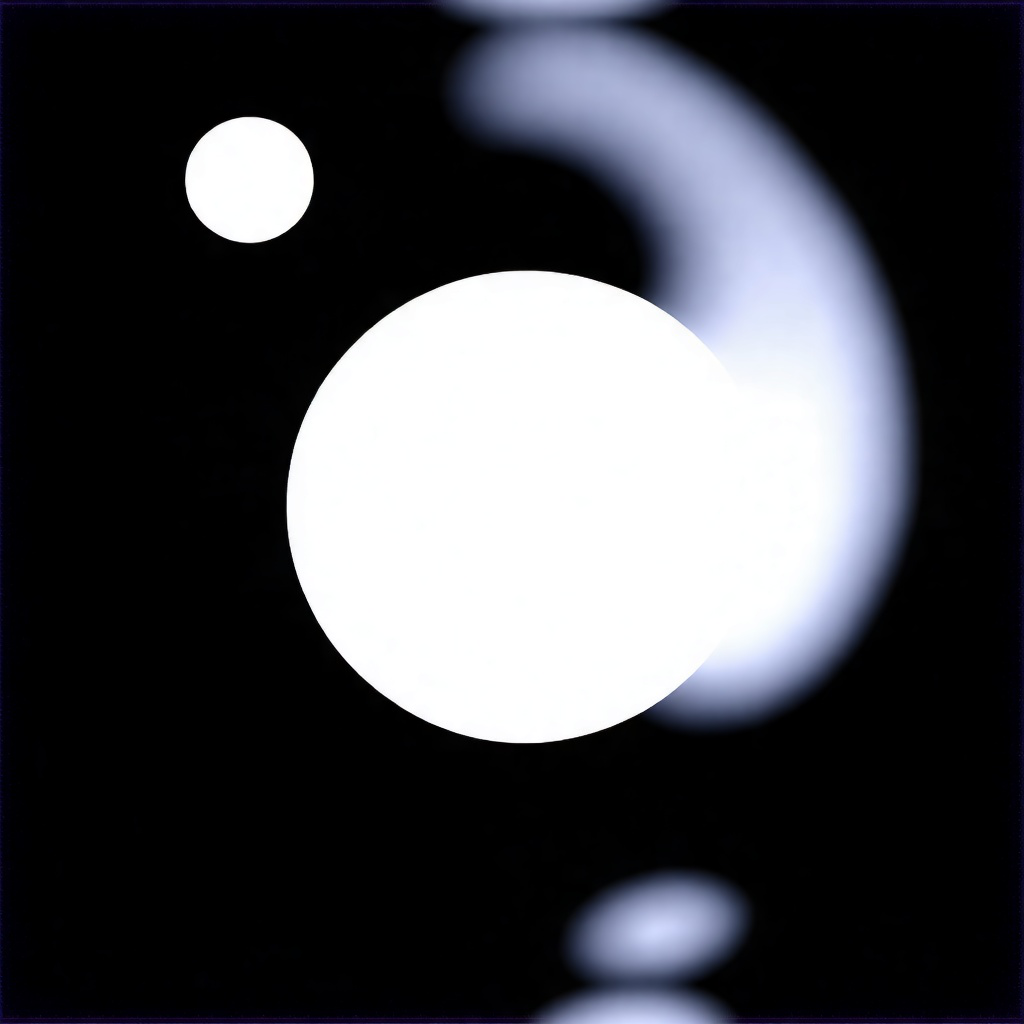}}
    \end{tabular}
    \caption{Sample images for prompt "circle" at 1024x1024 resolution}
    \label{fig:circle_res_1024x1024}
\end{figure}
\subsection{Prompt: Triangle}
Prompt: A black background with a small white solid triangle located in the top left corner. The triangle is oriented with its base parallel to the top edge of the image and its apex pointing downward.

\begin{figure}[htbp]
    \centering
    \begin{tabular}{@{}ccc@{}}
        \textbf{Stable Diffusion 3 Medium} & \textbf{Stable Diffusion 3.5 Large} & \textbf{Stable Diffusion 3.5 Large Turbo} \\
        \fbox{\includegraphics[width=0.28\textwidth]{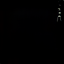}} &
        \fbox{\includegraphics[width=0.28\textwidth]{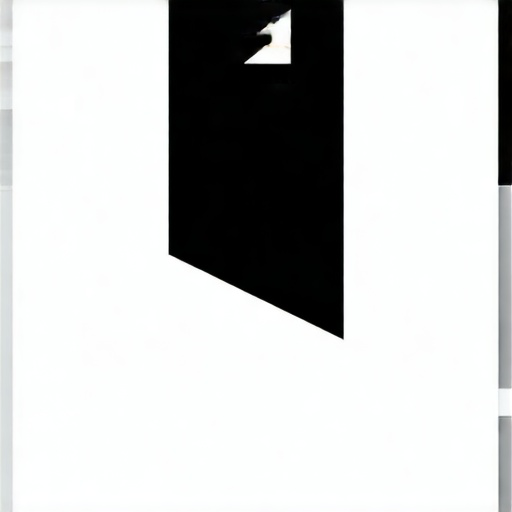}} &
        \fbox{\includegraphics[width=0.28\textwidth]{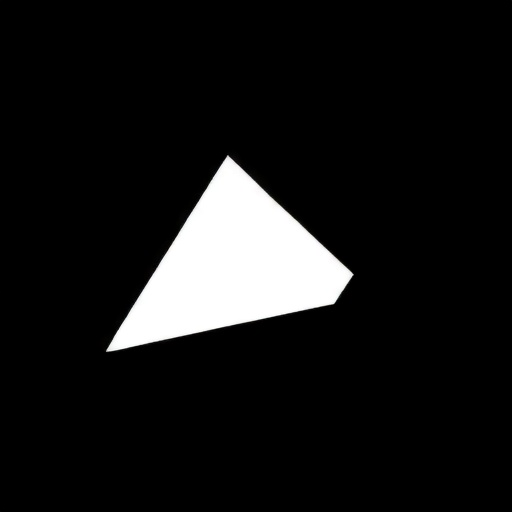}}
    \end{tabular}
    \caption{Sample images for prompt "triangle" at 64x64 resolution}
    \label{fig:triangle_res_64x64}
\end{figure}

\begin{figure}[htbp]
    \centering
    \begin{tabular}{@{}ccc@{}}
        \textbf{Stable Diffusion 3 Medium} & \textbf{Stable Diffusion 3.5 Large} & \textbf{Stable Diffusion 3.5 Large Turbo} \\
        \fbox{\includegraphics[width=0.28\textwidth]{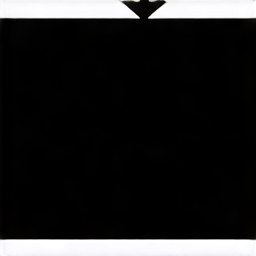}} &
        \fbox{\includegraphics[width=0.28\textwidth]{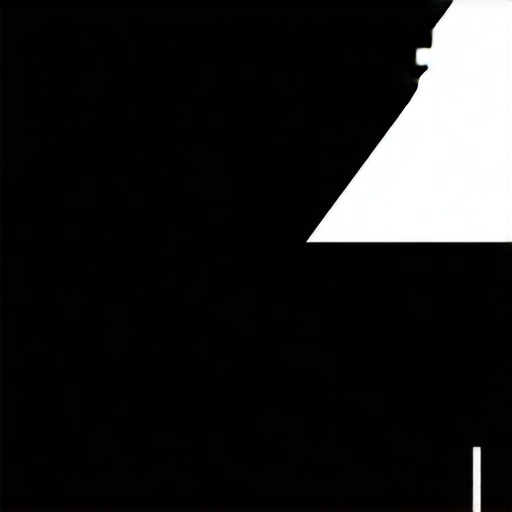}} &
        \fbox{\includegraphics[width=0.28\textwidth]{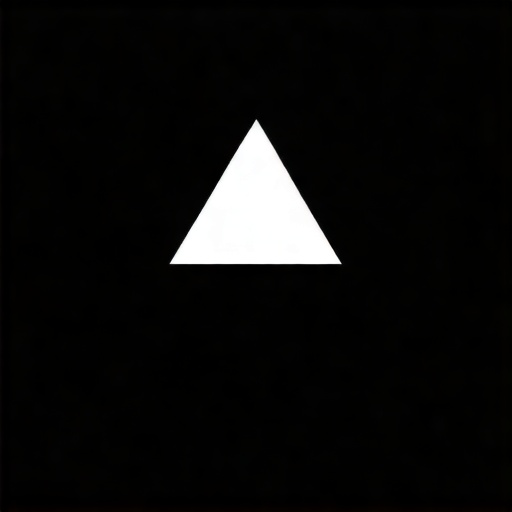}}
    \end{tabular}
    \caption{Sample images for prompt "triangle" at 256x256 resolution}
    \label{fig:triangle_res_256x256}
\end{figure}

\begin{figure}[htbp]
    \centering
    \begin{tabular}{@{}ccc@{}}
        \textbf{Stable Diffusion 3 Medium} & \textbf{Stable Diffusion 3.5 Large} & \textbf{Stable Diffusion 3.5 Large Turbo} \\
        \fbox{\includegraphics[width=0.28\textwidth]{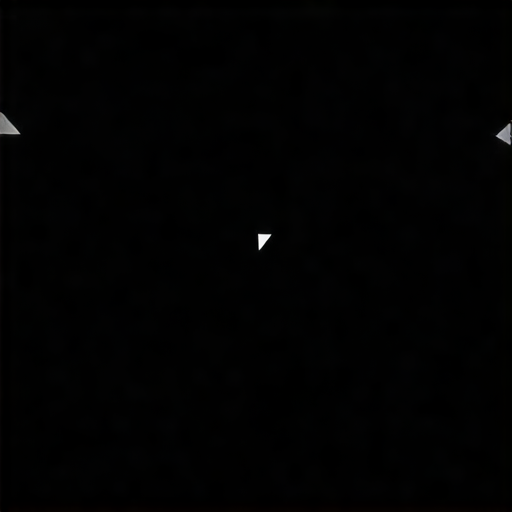}} &
        \fbox{\includegraphics[width=0.28\textwidth]{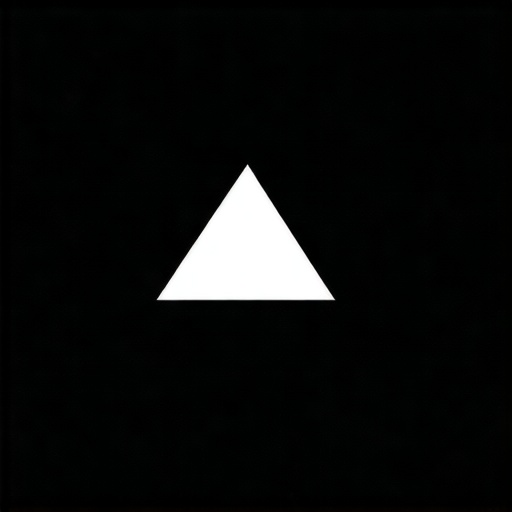}} &
        \fbox{\includegraphics[width=0.28\textwidth]{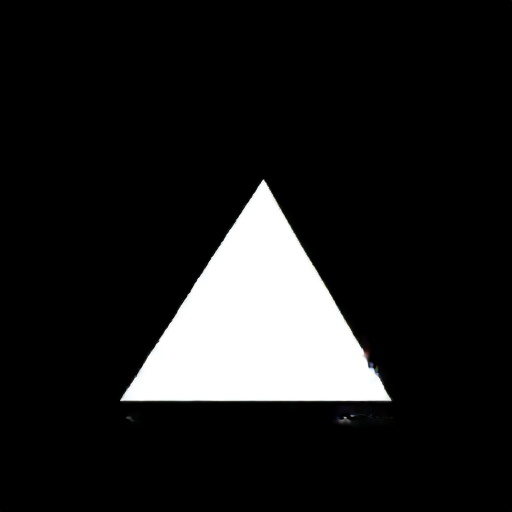}}
    \end{tabular}
    \caption{Sample images for prompt "triangle" at 512x512 resolution}
    \label{fig:triangle_res_512x512}
\end{figure}

\begin{figure}[H]
    \centering
    \begin{tabular}{@{}ccc@{}}
        \textbf{Stable Diffusion 3 Medium} & \textbf{Stable Diffusion 3.5 Large} & \textbf{Stable Diffusion 3.5 Large Turbo} \\
        \fbox{\includegraphics[width=0.28\textwidth]{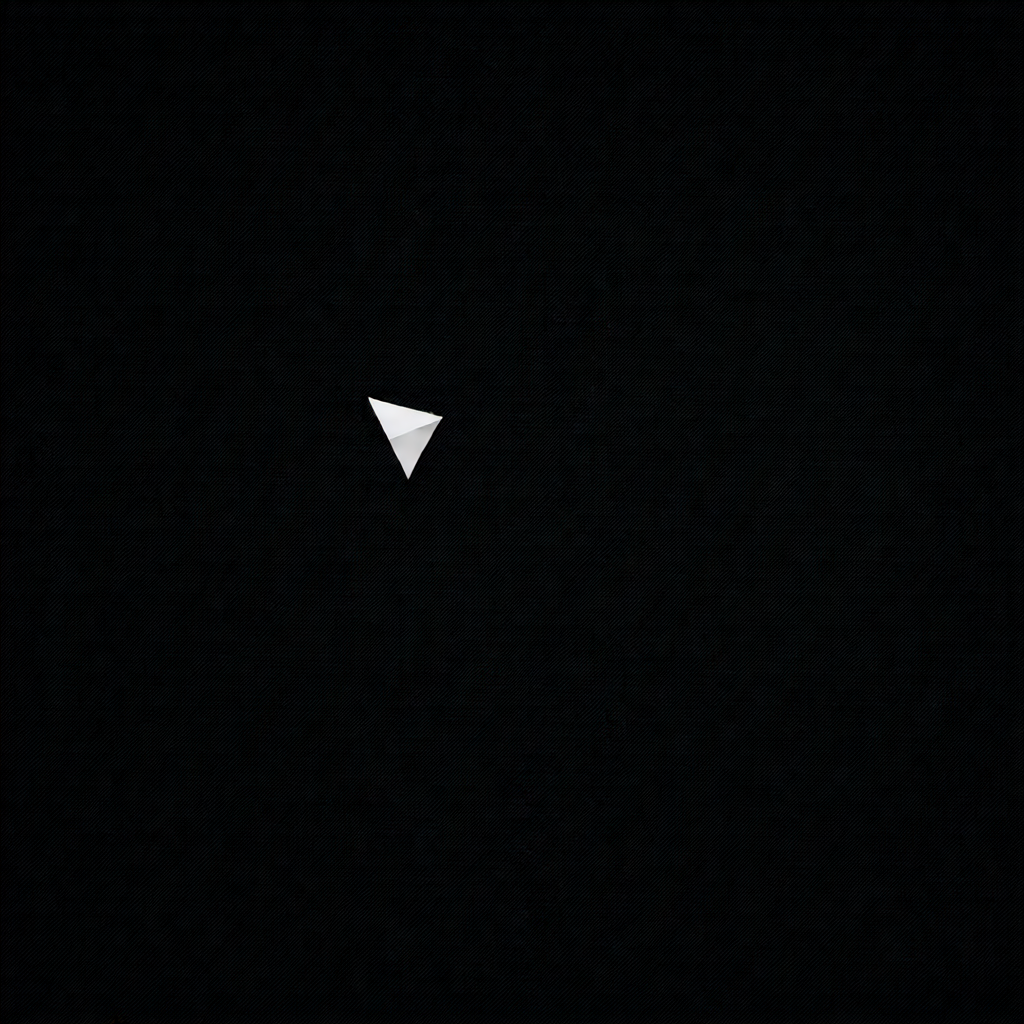}} &
        \fbox{\includegraphics[width=0.28\textwidth]{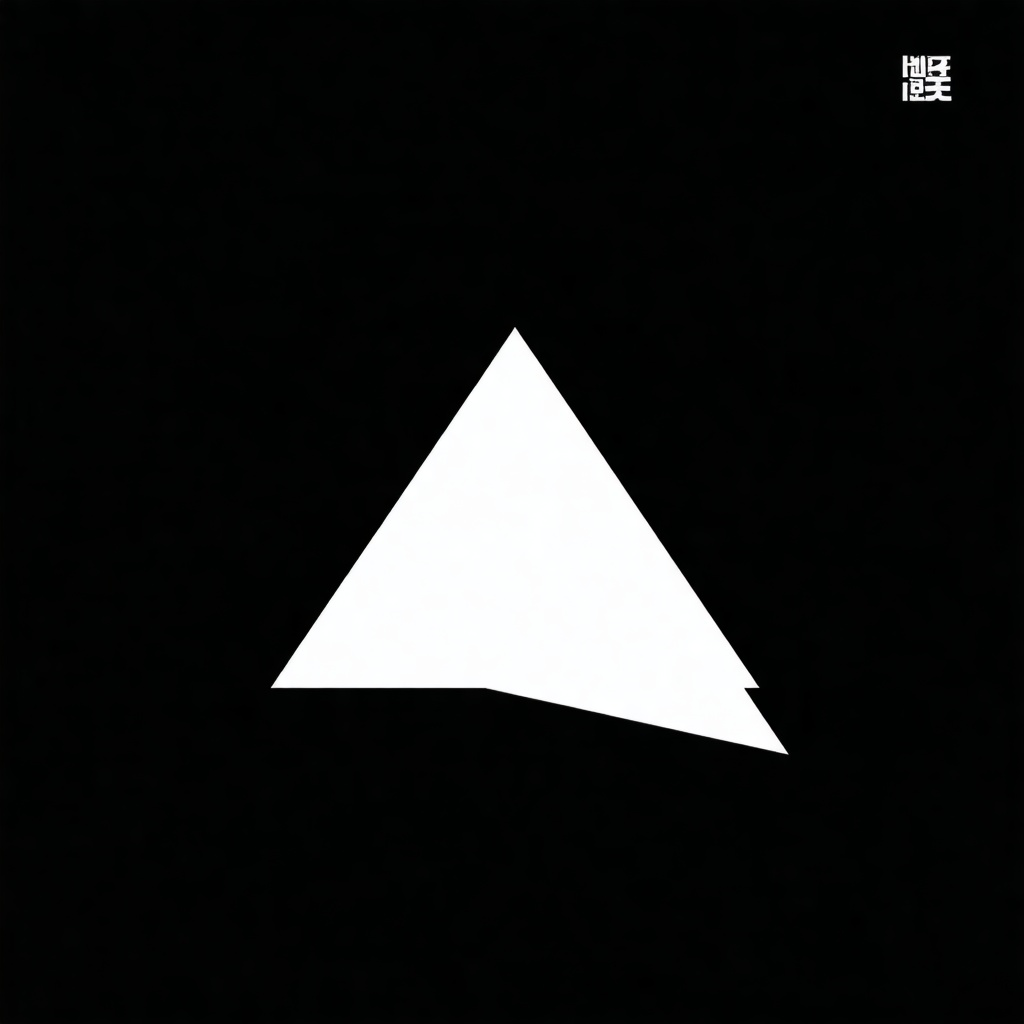}} &
        \fbox{\includegraphics[width=0.28\textwidth]{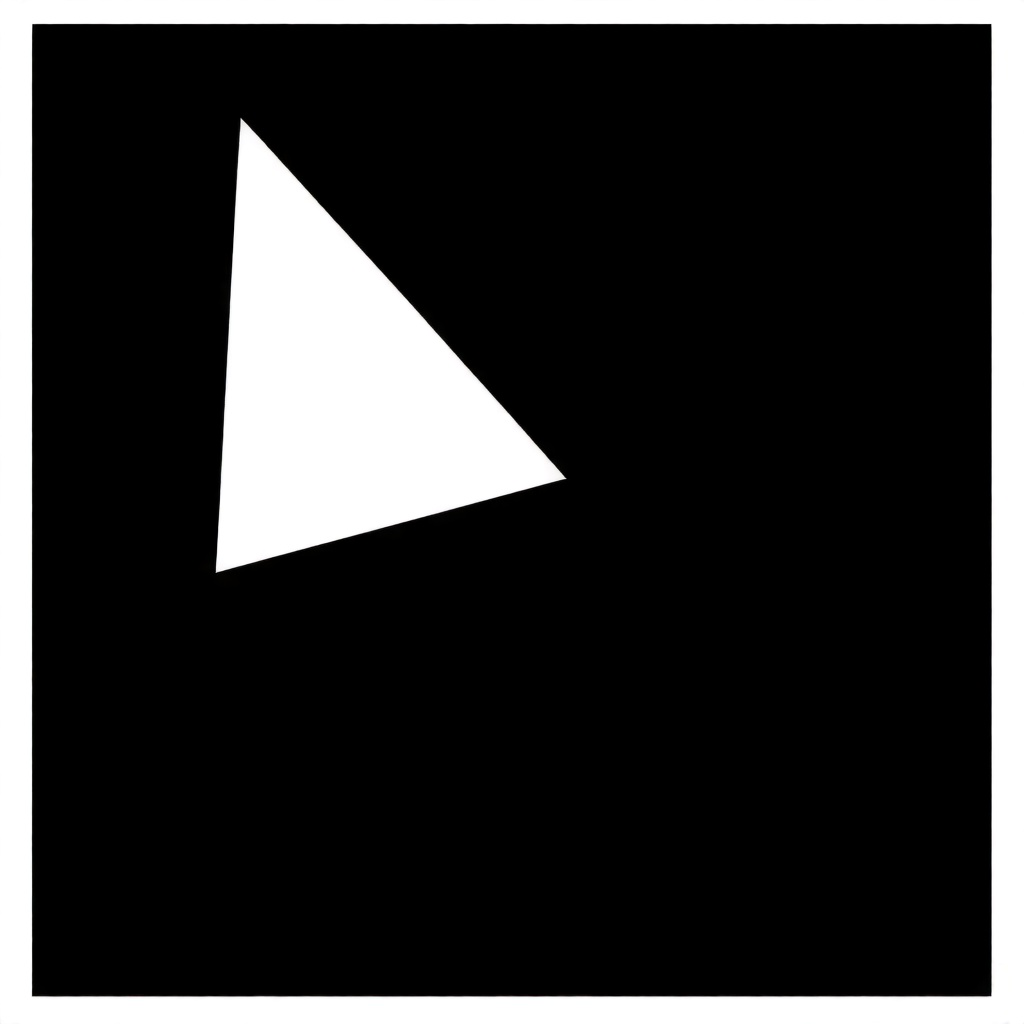}}
    \end{tabular}
    \caption{Sample images for prompt "triangle" at 1024x1024 resolution}
    \label{fig:triangle_res_1024x1024}
\end{figure}

\newpage

\section{Appendix B: Configuration}\label{sec:appendix_configuration}
\subsection{Prompts}
The following prompt is used with the \texttt{gpt-4o} model to generate the description of images that includes the shape and quadrant in which the center of the shape lies:
\begin{tcolorbox}[colback=gray!10,colframe=black,title=Prompt]
Analyse the image provided to you and provide the following three answers in the JSON format specified before (Only return the JSON response and strictly do not include any other comments or text):

1) Give a detailed description of the image which can be used as a text prompt to generate this image using any text to image model. Include features like background and foreground, main object of the image and their positioning and orientation, solid fill or outline etc among other important details to recreate this image. Be descriptive enough to recreate this image as faithfully as possible but limit it to 60 words, whichever is higher.

2) Provide the following encoding of shapes if a geometric shape is present as main object of the image with the following encoding:
- S for Square or Rectangle (only if the image contains a complete square or rectangle shape as the main object, output O if the shape is partial)
- C for Circle (only if the image contains a complete circle shape as the main object, output O if the shape is partial)
- T for Triangle (only if the image contains a complete triangle shape as the main object, output O if the shape is partial)
- O for Other cases not fitting into criteria of S, C and T

3) Provide the position of the center point of the main foreground object identified in the image with the following encoding:
- TL for Top Left quadrant (If your answer to question 2 is either S or C or T and the center of that shape falls in top left quadrant)
- TR for Top Right quadrant (If your answer to question 2 is either S or C or T and the center of that shape falls in top right quadrant)
- BL for Bottom Left quadrant (If your answer to question 2 is either S or C or T and the center of that shape falls in bottom left quadrant)
- BR for Bottom Right quadrant (If your answer to question 2 is either S or C or T and the center of that shape falls in bottom right quadrant)
- O for all Other cases

Do not make up an answer that fits the examples given to you, use the option "Other" given to you in the encoding if shape and quadrants don't match the criteria mentioned.

\end{tcolorbox}
\subsection{\textbf{Hardware}}
The section describes the hardware requirements to conduct our experiments successfully. Following are the GPU/CPU/OS/RAM details for training different iterations of VAE models, generating text descriptions with \texttt{gpt-4o}, and generating images with 3 different Stable Diffusion models for our dataset. Note that we have used multiple machines for parallel computing.

\begin{enumerate}
    \item NVIDIA A6000 workstation
    \begin{enumerate}
        \item Intel Xeon w9-3475X 72 core processor
        \item Nvidia 6000 Ada with 50GB VRAM
        \item 256GB RAM
        \item Operating System - Linux
    \end{enumerate}
    \item 3 NVIDIA H100 instances
    \item Pytorch version - 2.2.0
    \item Python version - 3.11.5
    \item CUDA version 12.1
\end{enumerate}
\subsection{\textbf{Factors of Variation}}
The dataset was configured to have 6 factors of variation. A combination of these 6 factors can completely define each image. Out of these 6 factors, only 4 vary in value: shape, scale, x co-ordinate and y co-ordinate of the center of the shape. Color and orientation are single-valued attributes across the dataset for reasons mentioned in \ref{sec:dataset}. All VAE models which have been trained have an 8 or 10 dimensional latent space as purely disentangled latent space is highly unlikely.
Following are the details regarding each factor of variation. We reiterate here from \ref{sec:dataset} that the center co-ordinates have been given enough margin from the center lines to mitigate confusion in quadrant assignment by \texttt{gpt-4o}o while generating the text descriptions.

\begin{enumerate}
    \item Shape: [0,1,2] corresponding to Square, Circle and Triangle.
    \item Scale: [0.5,0.75,1].
    \item X co-ordinate: 32 values starting from 25 to 205 with a jump of 5. Each value represents the exact pixel value in the 256 x 256 image.
    \item Y co-ordinate: Same as the X co-ordinate.
    \item Color: Only 1 color scheme. White shape on a black background.
    \item Orientation: Only 1 orientation with 0 degree tilt.
\end{enumerate}
Every possible combination of these factors lead to a 9216 image test dataset.
\subsection{\textbf{Binarization of Images}}
After transforms our dataset consists of binary grayscale images, i.e, each pixel is 0 or 1. However, the images generated by the generative models have floating point values between 0 and 1 which can skew the reconstruction loss. To cover all possibilities, we have done the inference using different thresholding values ranging from 160 to 250. We have also compared results with binarization against results without binarization. We conclude that thresholding does not affect our results in any noticeable way. Note that binarization sometimes acts negatively as far as reconstruction is concerned and can corrupt a well reconstructed image. However, such occurrences are meagre and do not affect our evaluation pipeline. 
\section{Appendix C: VAE}\label{sec:appendix_vae}
In the last decade, numerous adaptations of the VAE model have been developed to address the shortcomings of the conventional VAE model. Achieving an ideal output necessitates balancing the trade-off between minimizing reconstruction loss and KL divergence loss. $\beta$-VAE, $\beta$-TCVAE, among others, concentrate on improving disentanglement among the determinants of variation. The configuration of hyperparameters results in an extensive set; evaluating all possibilities necessitates considerable time and computational resources without assurance of success. We have conducted extensive research on related works using the dSprites dataset to identify the ideal values for each VAE hyperparameter and have implemented them accordingly. This section will provide a overview of all experiments conducted with and other information related to the VAEs.
\subsection{\textbf{VAE Architecture}}
We have used a typical ResNet architecture with skip connections for both our Encoder and Decoder networks. The VAE takes in a uni-channel image of size 256x256.
\subsubsection{\textbf{Encoder}}
The encoder comprises 8 convolutional ResNet blocks. Each block comprises two convolutional layers and a single skip connection. We configure the kernel size, stride, and padding to reduce the resolution of the input image by half after each block. We employed BatchNormalization and Parametric ReLU to mitigate overfitting and vanishing gradients. Ultimately, we obtain the tensors for mean and logarithmic variance
\subsubsection{\textbf{Decoder}}
We apply the same strategy as the encoder and transpose it to finally get a uni-channel image with the original shape of (256,256). The final activation layer is a sigmoid that produces the image with pixel values between 0 and 1 whereas the original image is a binary image.
\subsubsection{\textbf{VAE parameters}}
For the ${\beta}$-VAE we have used ${\beta}$ = 4  with the KL divergence term. We decided to not pursue the controlled capacity loss route as it is only required when the trade-off leans too much in favour towards disentanglement. However ${\beta}$-VAE gave a good enough reconstruction. As for ${\beta}$-TCVAE, the authors have stated that ${\alpha}$ = 1, ${\beta}$ = 6 and ${\gamma}$ = 1 gave the best results.
\subsubsection{\textbf{Reconstruction Loss}}

We have taken 2 different data distribution possibilities P(x) for our data. Separate training iterations have been carried out with respect to each assumption. Each assumption translates to a different formula for reconstruction loss.
\begin{enumerate}
    \item Bernoulli: translates to binary cross entropy loss between the reconstructed images and the original dataset images.
    \item Gaussian: translates to Mean Squared error loss between the reconstructed images and the original dataset images
\end{enumerate}
We note that for the Bernoulli distribution assumption, we could not get feasible results which led to us only working with the Gaussian assumption.
\subsubsection{\textbf{KL Divergence Loss}}
As we have assumed the latent space distribution to be normal, the KL Divergence term reduces to:
\begin{align*}
 \frac{1}{2} \sum_{j=1}^J \left(1 + \log ((\sigma_j)^2) - (\mu_j)^2 - (\sigma_j)^2 \right)
\end{align*}
where $\mu_j$ and $\sigma_j$ are tensors of shape (batch size, total latent dimensions). As far as ${\beta}$-VAE and a vanilla VAE are considered, the loss calculation is pretty straightforward in Python.
The authors of ${\beta}$-TCVAE have broken the KL divergence into 3 terms as follows:
\begin{equation*}\label{eq:decomposition}
\begin{split}
\mathbb{E}_{p(n)} \biggl[\KL 
\bigl( q(z|n) || p(z) \bigr) \biggr]
= \underbrace{\KL \left(q(z,n)||q(z)p(n) \right)}_{\text{{i} Index-Code MI}} + \underbrace{\KL \bigl( q(z) || \prod_j q(z_j) \bigr)}_{\text{{ii} Total Correlation}} + \underbrace{\sum_j \KL \left( q(z_j) || p(z_j) \right)}_{\text{{iii} Dimension-wise KL}}
\end{split}
\end{equation*}
where $z_j$ denotes the $j$th dimension of the latent variable. The ablation experiments carried out by the authors reveal that tuning ${\beta}$ leads to the best results. As stated in 8.1.3, we have used ${\alpha}$=${\gamma}$=1 and ${\beta}$=6 to focus on the Total Correlation term.
All 3 terms can be easily calculated in Python. 
\newpage
\section{Appendix D: Qualitative Evaluation}\label{sec:qualitative_evaluation}
In this section we will share some interesting results derived from our qualitative evaluation phase.
\subsection{Latent Space}
To gather evidence that our VAEs have been trained well, we must fulfill certain criteria:

\begin{enumerate}
    \item Each latent dimension must adhere to a normal distribution. This can be readily accomplished with statistical tests. The Anderson-Darling test is a widely utilized approach that produces a confidence statistic for a dataset to be assessed against a distribution with specified parameters.
    \item  A one-to-one correspondence between each factor of variation and latent dimension. Latent traversal, which examines the impact of altering a single latent dimension on the generated images, is a conventional technique employed in various studies pertaining to representation learning to assess this characteristic.
\end{enumerate}

In this section we shall present an in-depth version of our findings and show an example with one of the VAEs that we trained. A ${\beta}$-TCVAE trained on a Gaussian 8-dimensional latent space assumption for 300 epochs will be used to display our results.

\subsubsection{Adherence to Normal distribution}
The Scipy library has a pre-implemented function to perform the Anderson-Darling test. The test is applicable to Extreme Value Type 1 distributions, including Normal, Exponential, Logistic, Weibull minimum, and Gumbel distributions. It produces a "Anderson statistic" together with the distribution parameters and provides a definitive assessment of whether the distribution accurately represents the data sample.
For instance, if a VAE is trained with an 8-dimensional latent space, the ${\mu}$ tensor would possess a shape of (9216, 8). A sample set for the AD test comprises the values within a singular latent dimension (i.e., sample size = 9216). Each latent dimension constitutes a sample set and is individually assessed for conformity to a normal distribution. We observed positive results across every VAE we trained. For example, a ${\beta}$-TCVAE trained on a Gaussian 8-dimensional latent space assumption for 300 epochs showed that \textbf{all 8} latent dimensions follow a normal distribution.
\subsubsection{\textbf{One to one mapping of FoV with latent space}}
The theory that limiting the capacity of encoding channels promotes disentanglement is well found. Different factors of variation contribute differently to the reconstruction log likelihood. This pressure encourages the model to assign capacity to the most contributing factors in the initial stages of training. Further, this leads to the VAE finding representational axes in the latent space which align with the factors of variation.
Latent traversal is a popular method to test and visualize this property. We performed latent traversal on all dimensions of all our trained VAEs to have a qualitative outlook of disentanglement. Ideally, all dimensions should follow a normal distribution and while most do, there remain some with certain irregularities. In our case, it was the range. All dimensions resemble a bell curve but for some dimensions, the standard deviation is much less than 1. Therefore, instead of a fixed range ([-3,3] for a Normal distribution), we used 1.5 times the minimum and maximum value in a dimension across a batch of 128 images as the range for our latent traversal.
We shall show some of the more meaningful traversals we could find across all VAEs. We can broadly divide the traversals into 3 categories:
\begin{enumerate}

\item Category 1: No noticeable changes. Dimensions belonging to this category do not capture anything meaningful in image generation.
\item Category 2: Linear changes - a perfect traversal. We observe only 1 factor of variation with change in values. Dimensions belonging to this category are completely disentangled.
\item Category 3 - Entangled changes - we observe multiple factors of variation with change in values. Dimensions in this category form sub-groups which encode a particular factor of variation.
\end{enumerate}
To show the affect of ${\beta}$, we shall show traversals for different types of VAEs as well. Note that 2 separate disentangled dimensions may encode the same factor of variation.

\section*{\textbf{Vanilla VAE}}
\begin{figure}[H]
    \centering
    \includegraphics[width=1\textwidth]{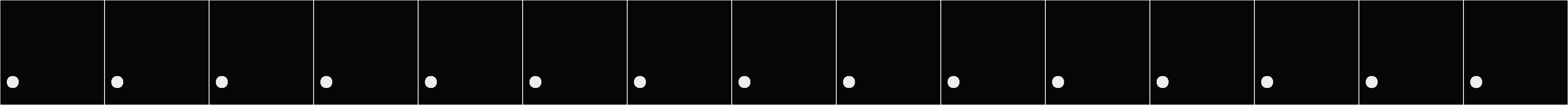} 
    \caption{Category 0 dimension}
    \label{fig:vvae-cat-0} 
\end{figure}

\begin{figure}[H]
    \centering
    \includegraphics[width=1\textwidth]{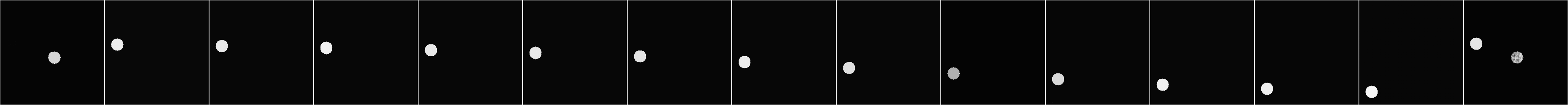} 
    \caption{Category 1 dimension}
    \label{fig:vvae-cat-1} 
\end{figure}

\begin{figure}[H]
    \centering
    \includegraphics[width=1\textwidth]{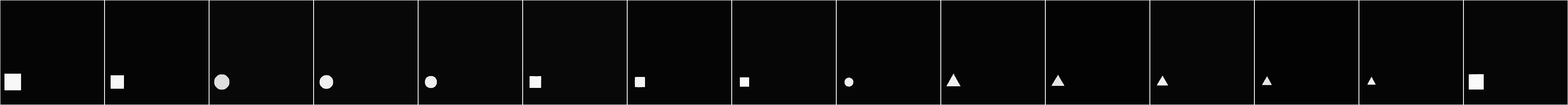} 
    \caption{Category 2 dimension}
    \label{fig:vvae-cat-2} 
\end{figure}

\section*{\textbf{${\beta}$-TCVAE}}

\begin{figure}[H]
    \centering
    \includegraphics[width=1\textwidth]{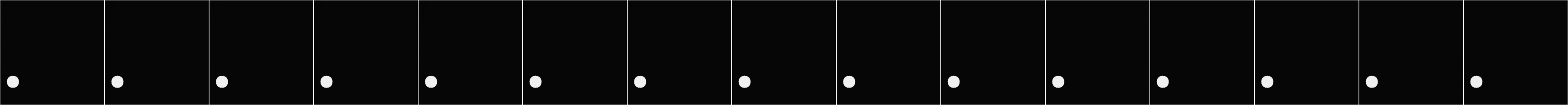} 
    \caption{Category 0 dimension}
    \label{fig:btcvae-cat-0} 
\end{figure}

\begin{figure}[H]
    \centering
    \includegraphics[width=1\textwidth]{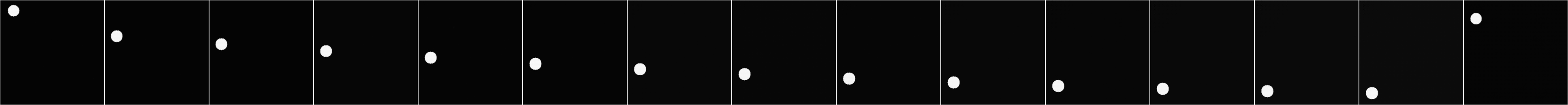} 
    \caption{Category 1 dimension}
    \label{fig:btcvae-cat-1} 
\end{figure}

\begin{figure}[H]
    \centering
    \includegraphics[width=1\textwidth]{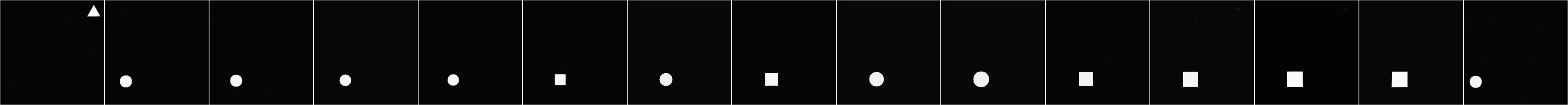} 
    \caption{Category 1 dimension}
    \label{fig:btcvae-cat-1-2} 
\end{figure}

\begin{figure}[H]
    \centering
    \includegraphics[width=1\textwidth]{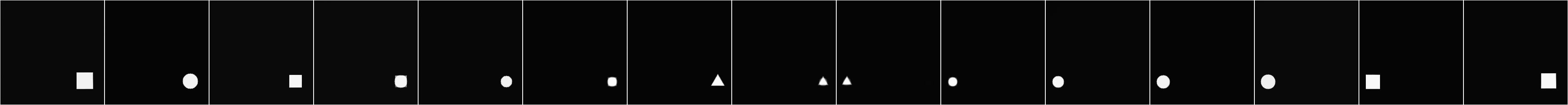} 
    \caption{Category 2 dimension}
    \label{fig:btcvae-cat-2} 
\end{figure}

\newpage
\section*{\textbf{${\beta}$-VAE}}
\begin{figure}[H]
    \centering
    \includegraphics[width=1\textwidth]{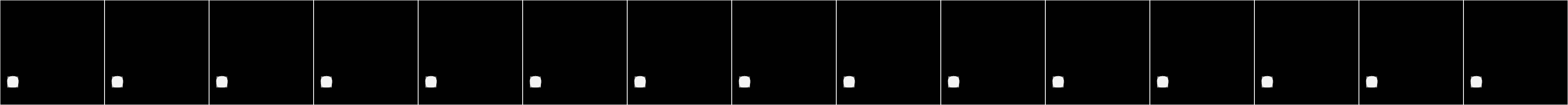} 
    \caption{Category 0 dimension}
    \label{fig:bvae-cat-0} 
\end{figure}

\begin{figure}[H]
    \centering
    \includegraphics[width=1\textwidth]{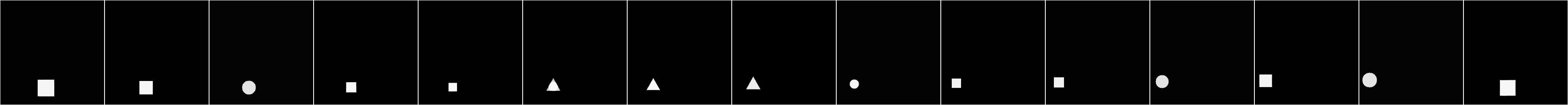} 
    \caption{Category 1 dimension}
    \label{fig:bvae-cat-1} 
\end{figure}

\begin{figure}[H]
    \centering
    \includegraphics[width=1\textwidth]{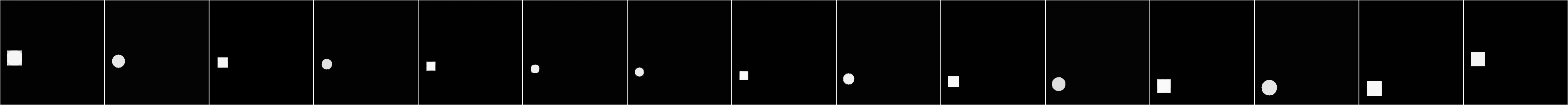} 
    \caption{Category 2 dimension}
    \label{fig:bvae-cat-2} 
\end{figure}

We would like to emphasize that even though we did not achieve pure disentanglement, our reconstruction loss section shows that our VAEs have learned the data distribution quite well and all factors of variation have been well documented.
\section{Appendix E: Additional Results}\label{sec:additional_results}
\subsection{\textbf{Impact of Threshold over Iterations}}\label{sec:Impact of Threshold over Iterations}
Here we reproduce the behavior of threshold having no effect on reconstruction of images with the rest of the trained VAEs.
\begin{enumerate}
    \item Vanilla VAE:
     \begin{figure}[H]
    \centering
    \includegraphics[width=1\textwidth]{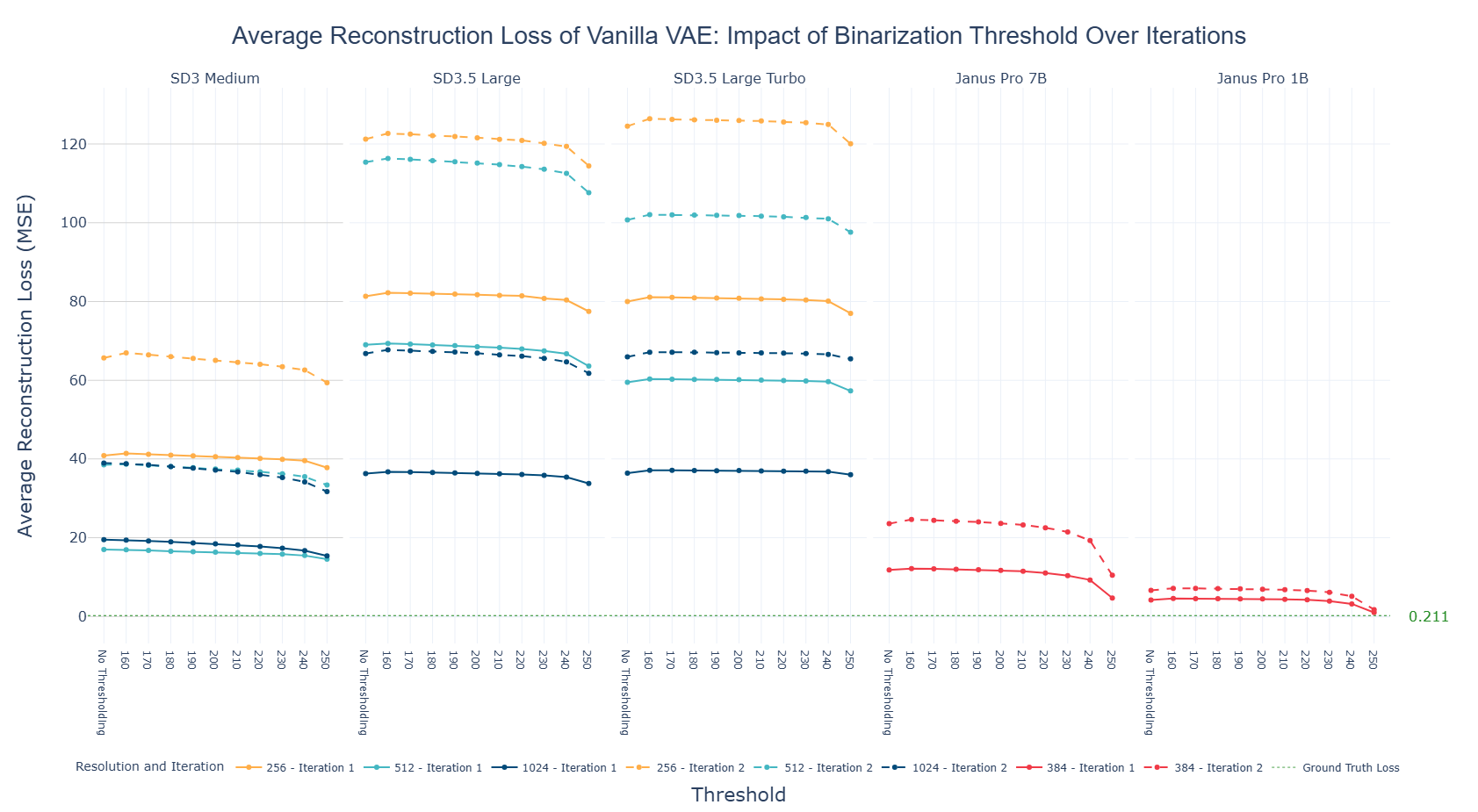}

    \caption{}
    \label{fig:appendix_figure_1.png} 
\end{figure}

    \item ${\beta}$-VAE: 
    \begin{figure}[H]
    \centering
    \includegraphics[width=1\textwidth]{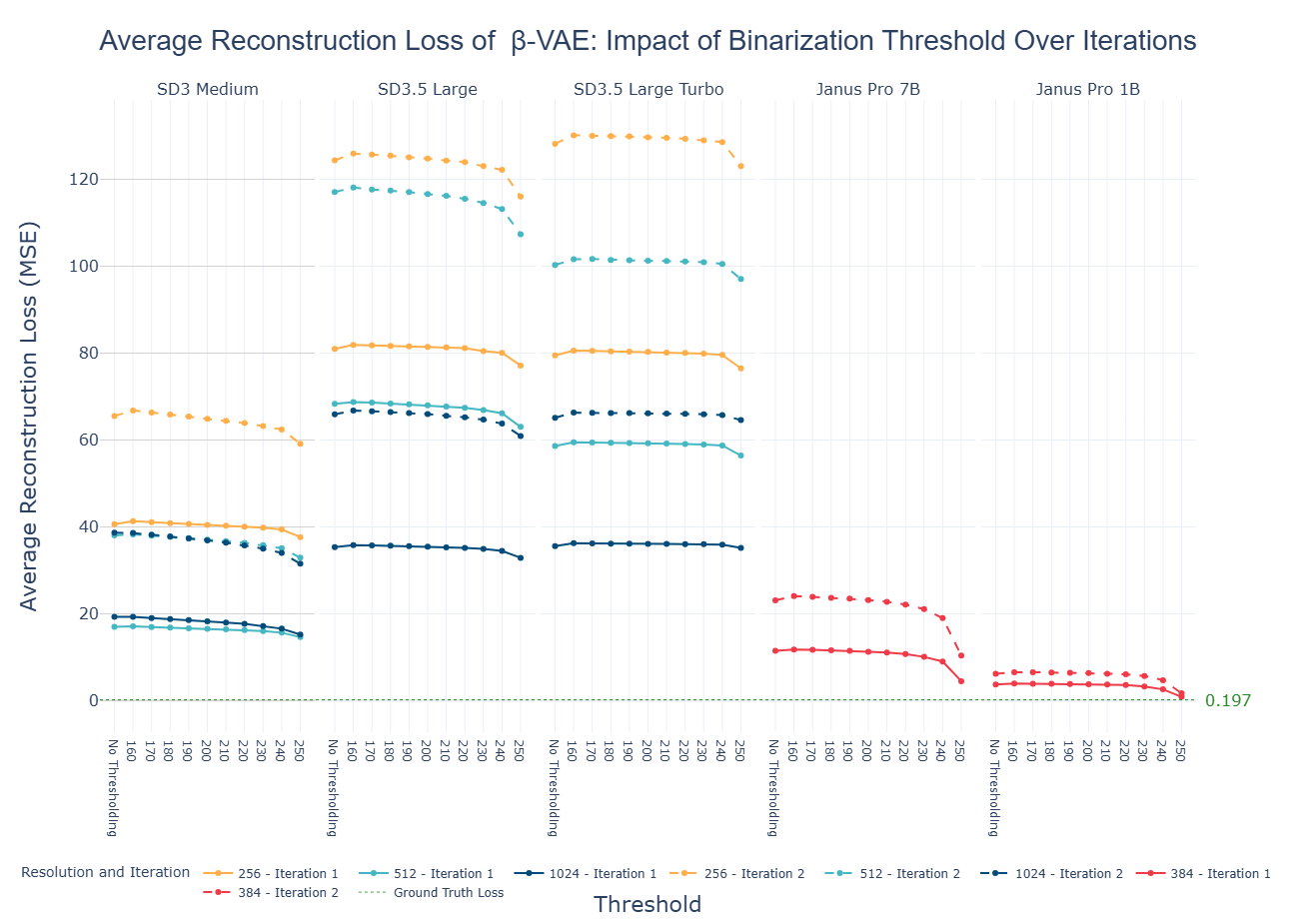}

    \caption{}
    \label{fig:appendix_figure_2.png} 
\end{figure}

\end{enumerate}
\subsection{\textbf{Impact of Resolution over Iterations}}\label{sec:Impact of Resolution over Iterations}
Here we see the behavior of the effect of resolution on reconstruction of images with the rest of the trained VAEs.
\begin{enumerate}
    \item Vanilla VAE:
     \begin{figure}[H]
    \centering
    \includegraphics[width=1\textwidth]{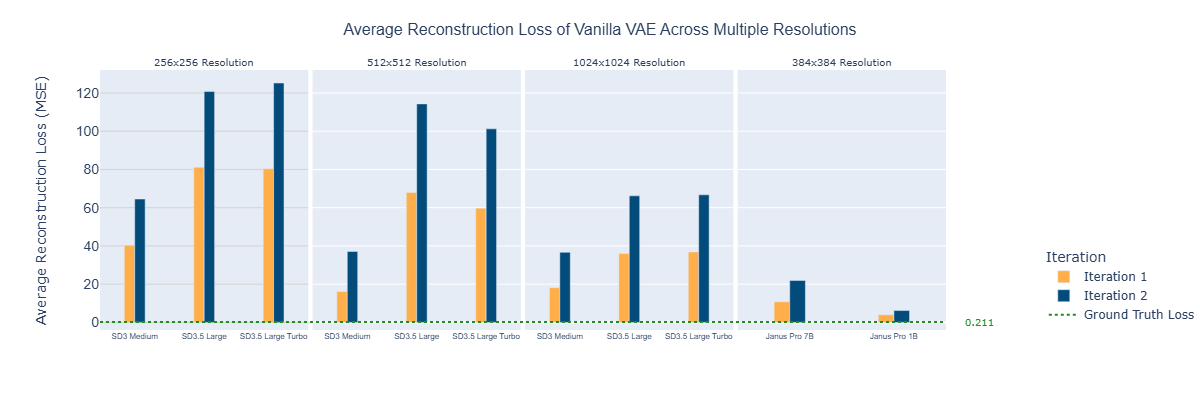}

    \caption{}
    \label{fig:appendix_figure_3.png} 
\end{figure}
    \item ${\beta}$-VAE: 
     \begin{figure}[H]
    \centering
    \includegraphics[width=1\textwidth]{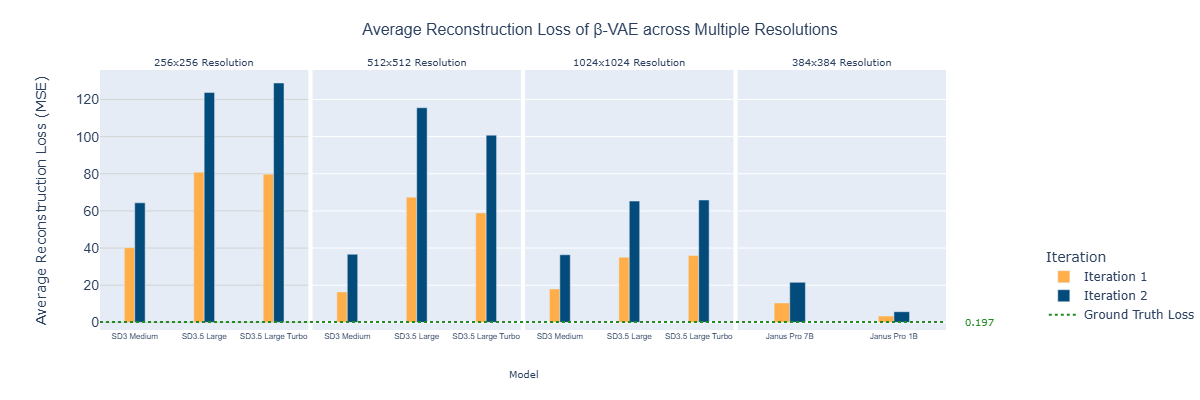}

    \caption{}
    \label{fig:appendix_figure_4.png} 
\end{figure}

\end{enumerate}

\end{document}